%% file: main.tex
\documentclass[lettersize,journal]{IEEEtran}
\usepackage{amsmath,amsfonts,amssymb}
\usepackage{algorithmic}
\usepackage{array}
\usepackage[caption=false,font=normalsize,labelfont=sf,textfont=sf]{subfig}
\usepackage{textcomp}
\usepackage{stfloats}
\usepackage{url}
\usepackage{graphicx}
\usepackage{booktabs}
\usepackage{xcolor}
\usepackage{cite}
\usepackage{bibunits}   
\usepackage{enumitem}
\usepackage{multirow}
\usepackage[subtle,tracking=normal]{savetrees}
\usepackage{xspace}

\newcommand{\ie}                {\emph{i.e.},\xspace}
\newcommand{\eg}                {\emph{e.g.},\xspace}
\newcommand{\etc}               {\emph{etc.}\xspace}

\newcommand{\naive}             {na\"{\i}ve\xspace}

\graphicspath{{figures/}{../../figures/}}

\makeatletter
\renewenvironment{IEEEbiographynophoto}[1]{%
\if@IEEEbiographyTOCentrynotmade%
  \setcounter{IEEEbiography}{-1}%
  \refstepcounter{IEEEbiography}%
  \addcontentsline{toc}{section}{Biographies}%
  \global\@IEEEbiographyTOCentrynotmadefalse%
\fi%
\refstepcounter{IEEEbiography}%
\addcontentsline{toc}{subsection}{#1}%
\normalfont\@IEEEcompsoconly{\sffamily}\footnotesize\interlinepenalty500%
\vskip 0.5\baselineskip plus 1fil minus 0\baselineskip%
\parskip=0pt\par%
\noindent\textbf{#1\ }\@IEEEgobbleleadPARNLSP}{\relax\par\normalfont}
\makeatother

\begin{document}

\title{Understanding Differentiable Embeddings Through\\ Differential and Integral Geometry}

\author{Xinyu~Zhang and Klaus~Mueller ~\IEEEmembership{Fellow,~IEEE}
\thanks{X.~Zhang is an independent researcher (e-mail: zhang146@cs.stonybrook.edu).}
\thanks{K.~Mueller is with the Department of Computer Science, Stony Brook
University, Stony Brook, NY, USA (e-mail: mueller@cs.stonybrook.edu).}
\thanks{(Corresponding author: Xinyu Zhang.)}
}

\markboth{IEEE Transactions on Visualization and Computer Graphics}%
{Author \MakeLowercase{\textit{et al.}}: Understanding Differentiable Embeddings Through Differential and Integral Geometry}

\maketitle

\begin{bibunit}[IEEEtran]   

\begin{abstract}
How can an analyst decide whether a nonlinear dimensionality-reduction (DR) embedding can be trusted? Existing diagnostics provide only partial answers: projection glyphs characterize local sensitivity, map-continuity scores measure local conditioning, and transport-based analyses reveal path-dependent inconsistencies. However, these methods appear unrelated and provide no common framework for understanding when they agree or disagree. We show that they are all derived from a single geometric object induced by every differentiable embedding, whether defined implicitly through optimization (e.g., t-SNE or UMAP) or explicitly by a learned mapping such as an autoencoder. This framework provides two complementary geometric views of an embedding. The differential view explains local behavior: its first-order term recovers projection glyphs, while its second-order curvature quantifies how far their linear approximation remains reliable. The integral view follows the same geometry along high-dimensional paths and determines whether an embedding depends only on the current state or also on the path taken to reach it. We further show that map-continuity is not itself a geometric reading of the embedding, but rather a prerequisite for the other analyses. The framework is theoretically complete for diagnostics derived from the embedding geometry, and we prove the integral view irreducible: no amount of local measurement at any number of points, to any order of derivative, reproduces what it detects. Classical rank-based metrics form a complementary class based on finite-scale neighborhood relationships. Experiments on synthetic and real datasets validate theoretical predictions, demonstrate accurate curvature-based trust estimates on single-cell embeddings (Spearman 0.963–0.999), and show that the integral analysis distinguishes single-valued embeddings from path-dependent optimization-based embeddings in ways that existing pointwise diagnostics cannot.
\end{abstract}


\section{Introduction}\label{sec:intro}
\IEEEPARstart{D}{}imensionality reduction (DR) has become one of the primary tools for exploring
high-dimensional (high-d) data: scientists routinely read two- or three-dimensional embeddings for clusters,
trajectories, transitions, and anomalies, from single-cell genomics to computer vision.
Every downstream interpretation depends on a single question:
\emph{Can an analyst trust what the embedding appears to say?}
A nonlinear embedding is neither globally faithful nor uniformly unreliable: some regions admit
meaningful local interpretations while others distort neighborhood relationships or apparent
trajectories. The practical problem is therefore rarely whether an embedding is ``good'' or ``bad'';
it is deciding \emph{when} a particular visual interpretation remains trustworthy.

\begin{figure*}[t]\centering
\includegraphics[width=\textwidth]{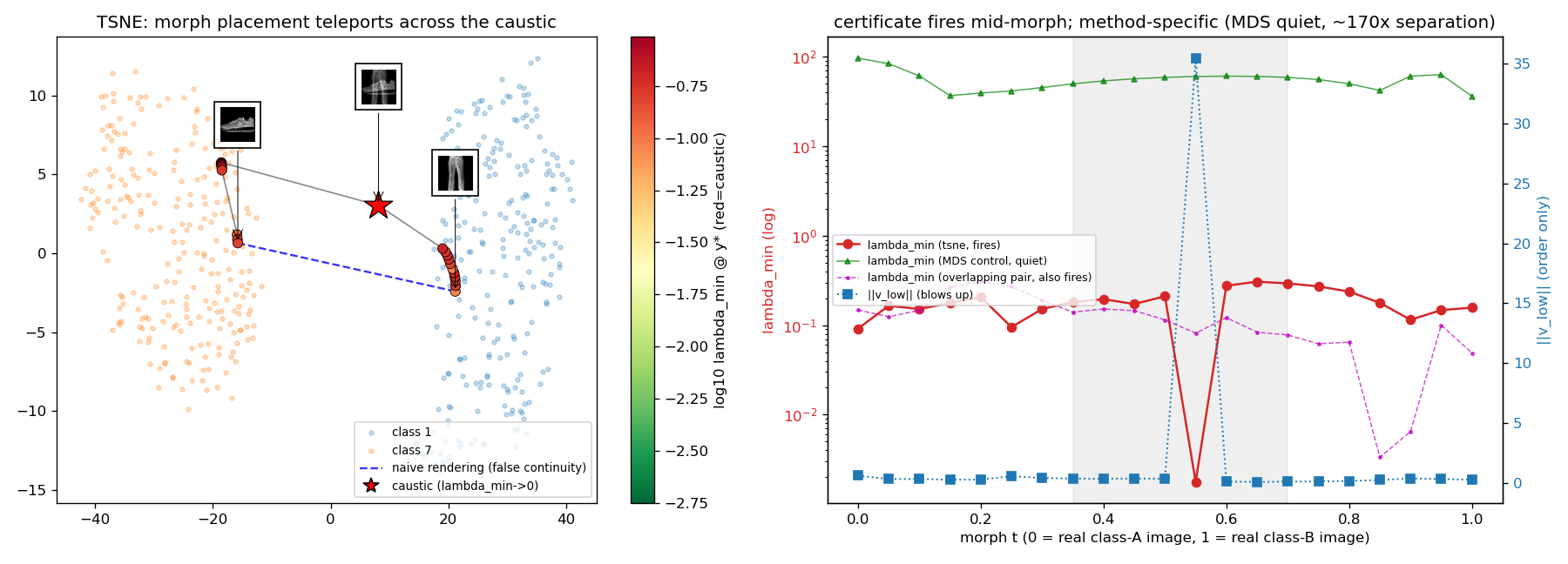}
\caption{\textbf{The induced flow makes an embedding's tears visible.} A pixel-space morph between a
FashionMNIST trouser and a sneaker, projected frame-by-frame into a fixed t-SNE map as
out-of-sample points.
\emph{Left:} the objective-consistent placement (dots coloured by $\log_{10}\lambda_{\min}$) does not
glide between the clusters; at the caustic $\Sigma=\{\det H_{yy}=0\}$ (red star,
$\lambda_{\min}\!\to\!0$) it \emph{teleports} across the gap, while the \naive straight interpolation
(dashed) asserts a smooth transition through a region the morph never occupies. \emph{Right:}
$\lambda_{\min}$ (log axis) collapses by orders of
magnitude at the crossing and the induced velocity $\|Jv\|$ spikes, as the spectral
divergence law predicts (Theorem~2); metric MDS on the same data stays
bounded away from zero ($168\times$ separation, green), and the
overlapping-class control (magenta) also fires.}
\label{fig:tear}
\end{figure*}

Existing diagnostic methods answer this question only partially: projection glyphs read local
sensitivity, map-continuity measures read local conditioning, and global transport
diagnostics~\cite{ref:conservative,ref:sevetlidis} expose path-dependent inconsistencies no local
check sees.
Each introduces its own mathematical object, visualization, and interpretation, leaving analysts without a unified language for reasoning about when local and global evidence agree or disagree.

Our move is to shift the unit of explanation from the individual diagnostics to the geometric object they read.
Every differentiable embedding induces a transport connection relating infinitesimal perturbations in the 
original space to motions on the embedding. For optimization-based DR the embedding is defined implicitly, as the 
minimizer of a conditional objective, and the connection follows by implicit differentiation of its optimality 
condition; for an explicit or parametric embedding (an autoencoder or parametric UMAP) the connection is simply 
the Jacobian of its map.

\begin{figure}[t]\centering
\includegraphics[width=0.8\columnwidth]{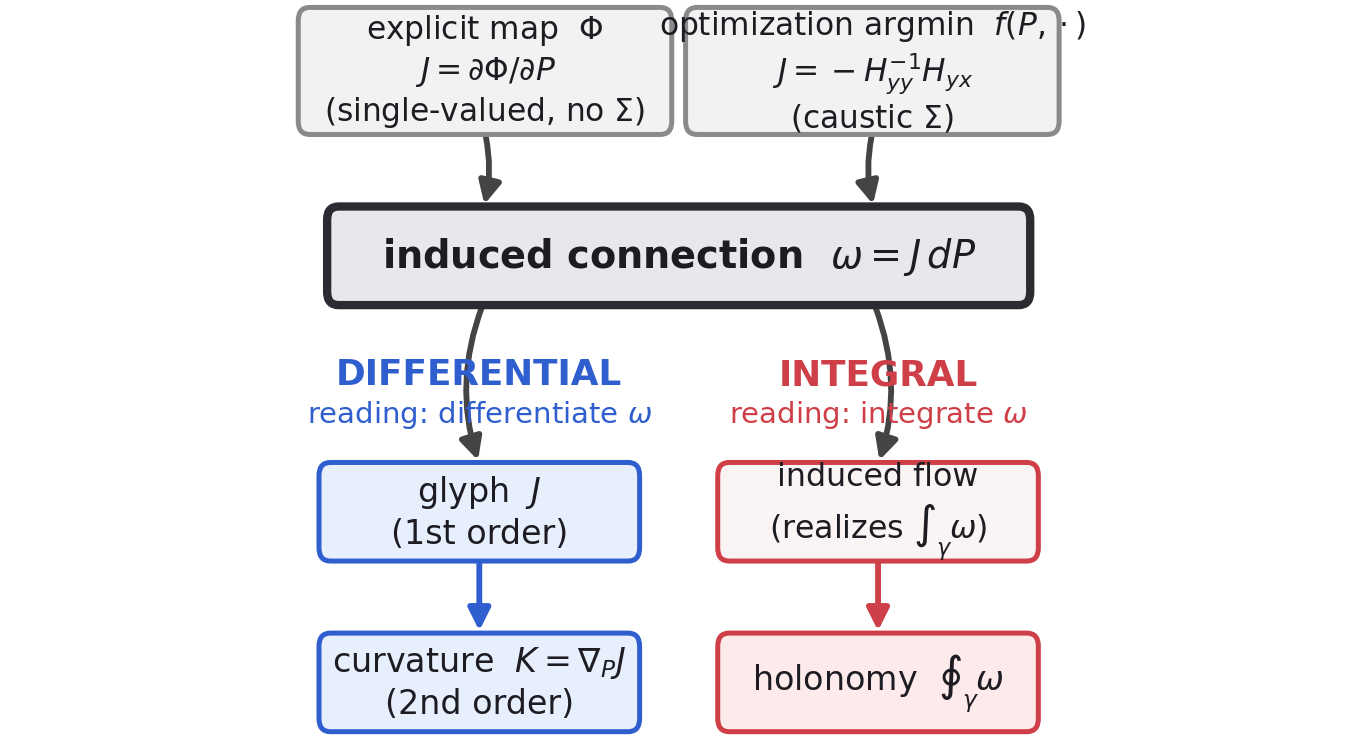}
\caption{One geometric object, two readings (schematic). Every differentiable embedding induces the
transport \emph{connection} $\omega=J\,dP$, whether its map is given explicitly or only through an
optimality condition; only the latter carries a caustic $\Sigma$, and only there does the integral
reading acquire content --- the \emph{holonomy} is non-zero exactly where the tracked branch fails to
cross $\Sigma$. The two readings hang off the Level-0 gate as siblings rather than as a chain:
neither is recoverable from the other (Cor.~1, Secs.~\ref{sec:operator}--\ref{sec:holonomy}).}
\label{fig:pipeline}
\end{figure}

Those readings come in two kinds. The \emph{differential} reading differentiates the
connection locally: its first order recovers the familiar projection glyph, its second introduces the
induced map's curvature as a certificate of how far that linear reading can be trusted. The
\emph{integral} reading accumulates the same object along finite paths and asks whether the embedding represents \emph{states}
rather than \emph{histories}, so that a placement does not depend on the route taken to reach it. Its
closed-loop case, the holonomy, is the one that carries a coordinate-free invariant. Projection
glyphs, curvature, and holonomy then appear as the successive differential orders and the integral
obstruction of one induced connection (Fig.~\ref{fig:pipeline}).

That account (the object, its two readings, the associated certificates) is the framework this paper contributes. Concretely, the paper makes three contributions:

\begin{itemize}
\item \emph{One geometry behind the existing diagnostics.} Projection glyphs, map-continuity
scores, and transport-based checks are not independent tools. Every differentiable embedding
induces a single geometric object, and we prove that any diagnostic reading that object is one of
its two views (Theorem~1, Sec.~\ref{sec:representation}). The result also reclassifies a diagnostic
in use: the map-continuity score is not a reading of the embedding but the precondition every
reading presupposes. Rescale the objective and the score moves while the embedding does not
(Sec.~\ref{sec:level0}).

\item \emph{Two views, provably independent.} The differential view certifies local reliability
through curvature; the integral view detects whether a placement depends on the route taken to
reach it. We prove them independent (Corollary~1): either can pass while the other fails, so a
reading falls into four regimes rather than onto a single quality axis. The consequential regime,
locally faithful yet route-dependent, is one no local check can signal (Sec.~\ref{sec:regimes}). A
second theorem ties both to the same conditioning quantity, so the existing score's divergence near
degeneracy is predicted rather than assumed (Theorem~2).

\item \emph{The framework put to work with a stated boundary.} Analysts most often read a point in
motion. Integrating the geometry traces a perturbation path through a fixed map continuously and at
amortized cost, marking where the embedding degenerates along the way (Sec.~\ref{sec:adv}).
Predictions are validated on synthetic data with analytic ground truth and on real single-cell and
image embeddings, where curvature-based trust estimates track measured error at Spearman
$0.963$--$0.999$. The classical rank-based metrics sit outside: the geometry controls when their
rankings flip but cannot report their\ref{sec:finitescale}).
\end{itemize}

Figure~\ref{fig:tear} is the framework read in one picture: carried through a fixed t-SNE map, a
high-d path does not glide between the clusters but \emph{teleports} across the gap where
the connection degenerates. Both readings speak there, and not as one check: the differential one bounds how far the cheap
local reading can be trusted; the integral one reports whether a single-valued placement exists,
independent certificates (Corollary~1) that here happen to fail together. Consequently, 
an analyst who trusted the dashed line reading would report a smooth
trouser-to-sneaker transition the morph never makes.

\section{Related Work}\label{sec:related}

Existing methods are complementary readings of projection geometry, or quality summaries of the
resulting embedding. We organize prior work by the geometric language of this paper and highlight 
the gap our framework addresses.

\subsection{Differential Readings of DR Embeddings}

Most existing projection diagnostics are differential: they characterize how infinitesimal
perturbations in the original space affect the embedding. Bian et al.~\cite{ref:bian} introduced
projection glyphs by implicitly differentiating local projection objectives, visualizing local
displacements directly on the embedding; Corbugy et al.~\cite{ref:corbugy} and Zabel et
al.~\cite{ref:zabel} extended this implicit-differentiation viewpoint to sensitivity analysis,
building on general implicit-differentiation formulations~\cite{ref:blondel,ref:lorraine}. Liu et
al.~\cite{ref:liu} quantified point-level map continuity through the smallest eigenvalue of the
embedding Hessian. DimReader~\cite{ref:dimreader}, pullback metrics~\cite{ref:pjm,ref:lso}, and
local length-distortion diagnostics~\cite{ref:aupetit,ref:checkviz} similarly characterize local
behavior.

In our terminology these are all differential readings of the same induced connection, and for the
distortion family the identification is concrete: the local length distortion along a direction $v$
is exactly $\|Jv\|$, so the per-point stretch these diagnostics visualize is governed by the singular
values of the induced Jacobian. Their characteristic pathology,
a short high-d segment drawn as a long embedding excursion, is precisely the divergence
$\sigma_{\max}(J)=\Theta(\lambda_{\min}^{-1})$ of our spectral divergence law
(Sec.~\ref{sec:spectral}). Our framework places them in a common language and makes the shared
representation carry more: the first-order $J=-H_{yy}^{-1}H_{yx}$ that Bian et al.\ render as a static
local-subspace glyph is here the connection itself, which also admits a second-order reading (the
induced map's curvature) and a path-level one, neither exposed by
any first-order diagnostic; a representation theorem (Sec.~\ref{sec:representation}) shows the
unification is a characterization rather than an analogy.
Curvature has separately been used to \emph{evaluate} embeddings and estimate intrinsic
dimension~\cite{ref:beylier}, and to \emph{drive} the embedding itself through graph-curvature edge
reweighting~\cite{ref:embedor}, but these measure the curvature of the data graph, whereas our
second-order reading measures the second fundamental form of the induced map.

The reverse direction that carries a low-dimensional (low-d) location back to high-d data
is a family of its own: built by local affine interpolation of the forward
projection~\cite{ref:ilamp}, learned outright as a network~\cite{ref:nninv}, and consumed
by applications such as classifier decision-boundary maps~\cite{ref:dbm}. Each fits an inverse as a
separate model; ours is the \emph{same} connection read backward, certified by the forward holonomy,
no separate estimators needed (Sec.~\ref{sec:inverse}).

\subsection{Integral Readings and Transport Consistency}

A separate line studies projection through transport, paths, or topology. Trajectory and dynamic embeddings~\cite{ref:dyntsne}, RNA-velocity
projections~\cite{ref:rnavelocity,ref:scvelo}, and transport-based visualization investigate how
structures evolve along finite paths; in machine learning, path consistency appears in studies of
conservative vector fields, autoencoder transport~\cite{ref:conservative}, gauge-invariant
representations~\cite{ref:sevetlidis}, holonomy, and related topological effects in
connection-based manifold learning~\cite{ref:ptu,ref:fibered}. These works typically begin from
externally defined vector fields, learned dynamics, or abstract geometric constructions; by contrast,
our integral reading is induced directly by the optimization geometry of the DR objective itself, so
holonomy emerges as a property of the induced connection rather than an independently introduced
diagnostic. 

The concurrent \emph{representation holonomy}~\cite{ref:sevetlidis} shares the name but transports
Procrustes-aligned activation subspaces of a DNN to compare models; ours transports the
out-of-sample (OOS) connection fixed by the DR objective's own optimality condition, and asks of a
\emph{single} embedding whether it represents states instead of histories.

\subsection{Projection Quality Assessment}\label{sec:quality}

A large body of visualization research evaluates embedding quality through summary measures: 
trustworthiness and continuity~\cite{ref:venna}, co-ranking and
rank-based neighborhood preservation~\cite{ref:coranking}, topological
preservation~\cite{ref:topoae}, inter-cluster reliability that localizes false and missing
groups~\cite{ref:steadiness}, and distortion measures, often aggregated over neighborhoods or the
whole dataset~\cite{ref:drsurvey}. A parallel thread builds interactive
tools that expose per-point projection error and let an analyst probe how the map
distorts~\cite{ref:probing,ref:tvisne}; that such checks matter in
practice is by now well documented~\cite{ref:stopmisusing}. Our goal is different: to explain
\emph{why} and \emph{where} an embedding becomes hard to interpret, complementing these summary
measures with pointwise geometric certificates.

\section{The Induced Connection}\label{sec:operator}
\subsection{From differentiable embedding to transport connection}
A differentiable embedding assigns to each high-d point $P$ a low-d position
through a map $\Phi$, and the object we read is its differential: a
\emph{transport connection} $\omega=J\,dP$, $J=d\Phi/dP$, the one-form that parallel-transports the
reading of a point along a high-d path. Two constructions realize $\Phi$, and both feed
the same connection.

\emph{Explicit embeddings.} An autoencoder, a parametric UMAP, 
or any learned
feature map gives $\Phi$ directly; the connection is its Jacobian $J=\partial\Phi/\partial P$, a
globally single-valued field, so $\omega=d\Phi$ is a one-form everywhere and its transport is
flat with holonomy identically zero. There is no caustic.

\emph{Optimization-based DR.} Metric MDS, t-SNE~\cite{ref:tsne}, UMAP~\cite{ref:umap,ref:umapjoss},
and Isomap~\cite{ref:isomap} \etc define $\Phi$ \emph{implicitly}. An OOS objective $f(P,y)$
scores how well a low-d position $y$ represents $P$ against a fixed set of anchors (the
conditional stress of MDS, the membership cross-entropy of t-SNE and UMAP, the geodesic discrepancy
of Isomap), and the embedding places $P$ at the conditional minimizer $y^\ast(P)=\arg\min_y f(P,y)$,
characterized by the optimality condition $\nabla_y f(P,y^\ast(P))=0$. This condition \emph{couples}
motion in $P$-space to motion in $y$-space: differentiating the identity
$\nabla_y f(P,y^\ast(P))=0$ in $P$ (the implicit function theorem) gives
$H_{yx}+H_{yy}\,dy^\ast/dP=0$, so we have the connection

\vspace{-10pt}
\begin{equation*}\label{eq:J}
J \;=\; \frac{dy^\ast}{dP} \;=\; -\,H_{yy}^{-1} H_{yx},
\qquad H_{yy}=\nabla^2_{yy} f,\; H_{yx}=\nabla^2_{yx} f.
\end{equation*}

Here $\Phi=y^\ast$ is single-valued only away from the caustic $\Sigma=\{\det H_{yy}=0\}$; the
object the rest of the paper reads is the connection $\omega$ itself.

\emph{The connection's transport is flat in
the interior, so all non-triviality of the integral reading comes from where the tracked branch fails
to extend across the caustic.} Wherever the induced map is smooth and single-valued --- everywhere
for an explicit embedding, and on the well-posed interior $P\setminus\Sigma$ for optimization-based
DR --- $\omega=d\Phi$ is an exact one-form, so interior transport is curvature-free and
path-independent, translating the reading by the period $\int_\gamma\omega$; its field strength
vanishes identically ($d\omega=0$) because $J=d_Py^\ast$ is a Jacobian, whose mixed partials commute,
and the connection is \emph{flat} and \emph{abelian} (its structure group is the translations
$(\mathbb R^d,+)$). Flatness is a
property of the transport instead of the map $y^\ast(P)$
which bends (Sec.~\ref{sec:curvature}).

We represent $J$ entirely by
automatic differentiation: the mixed term $H_{yx}v$ is a
Jacobian-vector product, and the $d\times d$ block $H_{yy}$ (with $d\in\{2,3\}$ the embedding
dimension) is a small dense Jacobian, regularized as $H_{yy}+\lambda_{\mathrm{reg}}I$ so that it
remains invertible near degeneracy; $\lambda_{\mathrm{reg}}=10^{-3}$ throughout unless stated.
The conditioning diagnostics of Sec.~\ref{sec:diff-scenario} are read at $\lambda_{\mathrm{reg}}=0$
so that no reported $\lambda_{\min}$ is floored by the regularizer.
The exactness propagates to both higher readings below.

\subsection{Level 0: the well-posedness gate}\label{sec:level0}
Before any reading can be taken, the connection has to exist. For an explicit map it always does, but for
the argmin construction it need not: the implicit function theorem delivers
$J=-H_{yy}^{-1}H_{yx}$ only where $H_{yy}$ is nonsingular, and the smallest eigenvalue
$\lambda_{\min}(H_{yy})$ measures how close the construction is to losing that precondition, its
vanishing locus being the caustic $\Sigma$ introduced above. This scalar is the quantity underlying the point-level
map-continuity score of Liu et al.~\cite{ref:liu}, which reports it as a singularity score diverging
as $\lambda_{\min}\to0$; the framework inherits it as the condition number of the
very problem it differentiates.

\smallskip\noindent\emph{The gate is upstream of the readings.} Replace $f$ by $cf$ with a constant
$c>0$: the minimizer $y^\ast(P)$ is
unchanged, hence so are the induced map $\Phi$ and its glyph
$J=-(cH_{yy})^{-1}(cH_{yx})=-H_{yy}^{-1}H_{yx}$, while $\lambda_{\min}(H_{yy})$ scales by $c$. A
quantity that moves while $\Phi$ stands still is not a function of $\Phi$, hence not a reading of the
induced map --- it fails the \emph{embedding-inducedness} by which every reading in this account is
defined (Sec.~\ref{sec:representation}). What $\lambda_{\min}$ reads is
the \emph{presentation} --- the conditioning of the optimality condition we differentiate --- not the
map that condition defines.

\smallskip
Every reading below presupposes the gate, and it resurfaces at each level in a different guise: it
sets the divergence rates of the differential magnitudes (Theorem~2, Sec.~\ref{sec:spectral}); it
localizes where an integrated path loses its branch, and licenses the \emph{uncertified} verdict of
the integral reading (Sec.~\ref{sec:holonomy}); it schedules where correction effort is spent along a
traced path (Sec.~\ref{sec:adv}); and its inverse-side counterpart is the rank deficiency of $J_T$, the
back-projection's own differential ill-posedness (Sec.~\ref{sec:inverse}). That reuse is the sense in
which the prior diagnostics are internally connected.

\subsection{Canonical representation and minimality}\label{sec:representation}
The framework's claim has two halves. Any \emph{natural} diagnostic (\ie one that only reads
the induced map's local or path-local geometry) must factor through the
connection's differential or integral readings; and none of the three levels it distinguishes can be
traded for the others. Naturality is three conditions, stated precisely in
App.~\ref{app:proofs}: the diagnostic is \emph{embedding-induced} (it sees the embedding only
through $\Phi=y^\ast$), \emph{local of finite order} (its value at $P$ depends only on the jet
$j^r_Py^\ast$) or \emph{path-local} (it sees $y^\ast$ along $\gamma$ only through the transport
$\dot y=J\dot\gamma$), and \emph{coordinate-equivariant} under Euclidean isometries of the embedding
plane.

\smallskip\noindent\textbf{Theorem~1 (canonical representation and minimality).} \emph{Let $\Phi$ be smooth and single-valued on the domain $U$
(trivially matching an explicit embedding, or the well-posed interior $P\setminus\Sigma$ for optimization-based DR). Then:
\begin{enumerate}[label=(\roman*)]
    \item Every natural first-order local diagnostic factors through the glyph $J=d_Py^\ast$; if scalar (Euclidean invariant), it factors through the pullback metric $J^\top J$.
    \item Every natural second-order diagnostic factors through $(J,\nabla J)$. If it is additionally affine-null (vanishing whenever $y^\ast$ is affine), its nontrivial content is carried entirely by the curvature $K=\nabla J$.
    \item Every natural loop-closure diagnostic factors through the induced holonomy $\mathrm{Hol}_\omega(\gamma)=\mathcal{T}^\omega_\gamma-I$ of $\omega=J\,dP$.
\end{enumerate}
Moreover the three levels are non-redundant, the integral one necessarily so:
\begin{enumerate}[label=(\roman*),start=4]
    \item \emph{(Minimality.)} The integral level cannot be traded for any finite collection of finite-order local readings. For any finite sample $S\subset\mathcal U_0$, any order $m$, and any loop $\ell$ avoiding $S$, there exist two smooth objectives whose induced branches agree to order $m$ at every point of $S$ --- so that every finite-order local reading returns identical values on $S$ --- while one closes the loop $\ell$ and the other does not. The loop holonomy is therefore the minimal integral obstruction to path independence.
\end{enumerate}
(See App.~\ref{app:proofs} for the naturality conditions and full proofs.)}

\smallskip
Naturality asks a diagnostic to be a function of a jet of $\Phi$, which excludes every diagnostic
carrying a length scale or a global fit. The dominant family of DR quality metrics, \eg rank- and
neighborhood-set-based scores such as trustworthiness, continuity~\cite{ref:venna}, and
co-ranking~\cite{ref:coranking}, is \emph{not} natural in this sense; it forms the complementary
class of Sec.~\ref{sec:quality}, wherein each metric reads the discrete embedding as a point set,
not the induced map's jets. Neither are several per-point diagnostics a practitioner would call
local: an axis-line scalar field fitted by least squares over all points~\cite{ref:dimreader},
proximity-distortion maps built from inter-point distances~\cite{ref:aupetit}, and
permutation-based per-point reliability scores~\cite{ref:scdeed} each determine their value at $P$
from data beyond any jet at $P$. Completeness here is about the natural class: inside it
the levels are fixed and irredundant, and the metrics outside it are outside for a stated structural
reason (Sec.~\ref{sec:finitescale}).

\smallskip
Clauses (i)--(iii) are statements about the well-posed interior: they presuppose a branch that is
smooth and single-valued on $U$, which is precisely the hypothesis that fails at the caustic where
the phenomena of Sec.~\ref{sec:integral} live. Clause (iv) is what reaches across. Its witnesses are
two genuine smooth objectives, agreeing to order $m$ at every sampled point yet differing in
branch-continuation on an unsampled arc (App.~\ref{app:proofs}), so the separation holds within the
DR-induced class the framework reads, not merely at the level of abstract one-forms: finite-order
local sampling is provably blind to an obstruction that embedding-induced transport genuinely
carries. The non-zero t-SNE holonomy of Sec.~\ref{sec:holonomy} instantiates this (P2); it persists
along loops where the local jets are unremarkable, so the integral reading is not a mechanical
integration of the glyph.

\smallskip\noindent\textbf{Corollary~1 (structural independence).} \emph{The map curvature and the
holonomy are functionally independent: no relation $\mathrm{Hol}=g(\mathrm{relK})$ can hold.
Structurally, the reading ladder (Sec.~\ref{sec:ladder}) exhibits the two possible zero patterns at
equal curvature sign: a strongly convex non-quadratic objective gives $\mathrm{relK}\neq0$ with
holonomy $\equiv0$, a non-convex one gives both non-zero. Empirically the witnesses are its two
explicit-map constructions: the autoencoder carries curvature of the same order as t-SNE
($\mathrm{relK}$ $1.5$--$4.6$ vs.\ $2.7$--$4.8$ across seeds) yet its holonomy $\ll$ t-SNE's. A
single-valued $g$ would force equal holonomy at equal curvature; it does not exist.}

\smallskip
The independence rests on the \emph{zero pattern} --- arbitrarily large curvature coexisting with an
exactly vanishing holonomy. That the holonomy is therefore \emph{not}
the loop-integral of the curvature we make constructive in App.~\ref{app:proofs} (Corollary~3).
The next two sections are the two branches the theorem leaves open, the differential and the integral
reading, respectively.


\section{The Differential Reading}\label{sec:differential}
\subsection{The spectral divergence law}\label{sec:spectral}
Theorem~1 fixes \emph{what} the readings are; before taking them one at a time, we record \emph{how}
they respond as the embedding approaches its degeneracy. That response is the framework's
quantitative half: the readings of this section and the next answer to a single spectral source, the
differential ones at predicted integer rates.

\smallskip\noindent\textbf{Theorem~2 (spectral divergence law).} \emph{The connection's primitive
first- and second-order generators inherit their singular behaviour from a single scalar, the
smallest Hessian eigenvalue $\lambda_{\min}(H_{yy})$ of the connection; a diagnostic derived from
them depends in addition on its own defining function, so what follows bounds the generators, not
every reading built on them. On the fold locus $D_{\mathrm{loc}}=\{\det H_{yy}=0\}$ the point-level score
diverges, $\lambda_{\min}^{-1}\to\infty$; approaching it, the first-order operator grows as
$\|J\|=\mathcal{O}(\lambda_{\min}^{-1})$ and the second-order curvature as $\|K\|=\mathcal{O}(\lambda_{\min}^{-3})$. The
upper bounds are attained when the soft mode is excited ($\|u^{\top}H_{yx}\|$
non-negligible for $u$ the $\lambda_{\min}$-eigenvector). (Proof in App.~\ref{app:proofs}.)}

\smallskip
Theorem~2's exponents are pointwise: the holonomy shares
the \emph{same source} --- it is non-zero only where the loop meets the caustic $\{\lambda_{\min}=0\}$
--- but registers it as a branch obstruction, not a $\lambda_{\min}$-power, so no divergence rate is
claimed for it.
Bian et al.'s glyph and Liu et al.'s $\lambda_{\min}^{-1}$ are the two \emph{exponent-one} 
members of a single derived ladder, while the second-order curvature forms its \emph{exponent-three} member. 
These rates follow from a single spectral source, so the divergence of Liu et al.'s singularity score
($\lambda_{\min}^{-1}\to\infty$) is predicted --- a claim about \emph{rates}, the score itself remaining the upstream gate of Sec.~\ref{sec:level0}. On controlled
synthetics the law holds to the predicted \emph{integer} exponents, as upper bounds
(Fig.~\ref{fig:rateladder}, App.~\ref{app:proofs}).
Metric MDS fits tightly ($0.88$ excitability) where t-SNE does not ($0.33$), leaving the
$\lambda_{\min}^{-3}$ bound loose on the latter, as the excitability condition predicts; which
methods excite the soft mode in general is not settled by the two measured here.

\smallskip\noindent\textbf{Proposition 3 (Local--global complementarity).} \emph{Let
$D=D_{\mathrm{loc}}\cup\{\text{argmin non-unique}\}$.
\emph{(i)} The pointwise conditioning magnitudes, including the glyph norm $\|J\|$, condition number 
$\lambda_{\min}^{-1}$, and curvature $\|K\|$, are driven by $\lambda_{\min}$ and stay bounded off $D_{\mathrm{loc}}$,
so at a basin reconfiguration in $\{\text{argmin non-unique}\}\setminus D_{\mathrm{loc}}$ 
where both competing branches remain nondegenerate, they register nothing. In contrast, 
the flow and holonomy (P2) actively detect the transition. In this sense, the reach of diverging differential magnitudes is 
strictly contained within that of the integral reading.
\emph{(ii)} The reverse containment fails: the interior integrability form $F$ (Cor.~3) is a purely local, second-order reading
that is non-zero on well-posed, integrable embeddings --- whose transport is flat (loop holonomy $\equiv0$) yet whose induced
map still curves ($K\neq0$) --- so the endpoint transport cannot recover it. Neither indicator family contains the other;
they are complementary detectors.}

\smallskip
The witnesses are, for~(i), an off-manifold basin-swap probe across which both competing branches
keep $\lambda_{\min}$ positive and $\|K\|$ bounded, and for~(ii), the interior form $F$ of Cor.~3 on
well-posed MDS. Numerically the pole is never reached: the $H_{yy}+\lambda_{\mathrm{reg}}I$
regularization saturates it at $\sim1/\lambda_{\mathrm{reg}}$, which is why we report the magnitudes
as saturating detectors.

\subsection{Level 1: the glyph}\label{sec:glyph}
\begin{figure}[t]\centering
\includegraphics[width=\columnwidth]{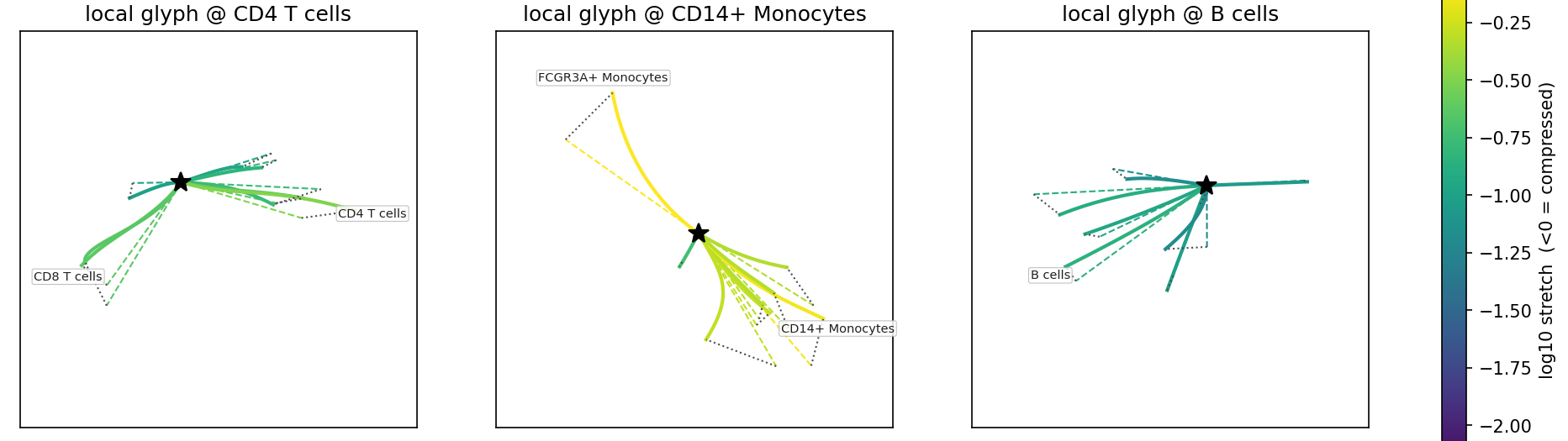}
\caption{The differential reading per cell on real data (PBMC3k, t-SNE). At each exemplar cell
(star), a common local high-d step toward each nearest-neighbour direction is read
three ways: the first-order \emph{glyph} (dashed straight arrow $y_0+Jd$), the \emph{integrated}
trajectory of the coupled flow (solid), and their \emph{gap} (dotted), the second-order
curvature term. The star of arms exposes directional \emph{anisotropy}; colour encodes
each direction's log stretch ($<0$ compressed). Arm-end labels name the neighbour cell type that
defines each input direction.}
\label{fig:localglyph}
\end{figure}

Applied to a basis of the local subspace, $Jv$ is the instantaneous
motion of $y^\ast$ under an infinitesimal push of $P$ along $v$. Collected over the basis, these
velocities are the linear transform that renders as static glyphs in~\cite{ref:bian}. 
In our framework the glyph can
be \emph{integrated}: pushing $P$ along a finite high-d path and integrating the
coupled flow of $y^\ast$ produces a trajectory whose initial tangent is the glyph. A
germ-convergence experiment confirms that the two are readings of one field: the finite-difference
chord of the integrated trajectory converges to the glyph column as the step vanishes, at
convergence order $\approx0.98$ on a swiss-roll/Isomap embedding and to machine precision on a
plane/MDS one. The static glyph and the dynamic trajectory are thus the same induced object read at
two scales. Figure~\ref{fig:localglyph} shows both readings together per cell on real data
(PBMC3k): along each named-neighbour direction the glyph and the integrated trajectory diverge
exactly where curvature sets in.

\begin{figure}[t]\centering
\includegraphics[width=0.6\columnwidth]{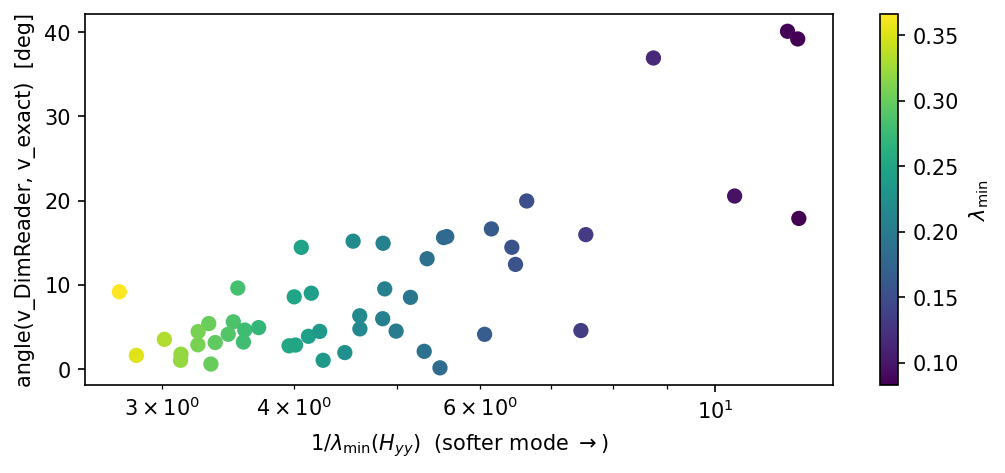}
\includegraphics[width=0.35\columnwidth]{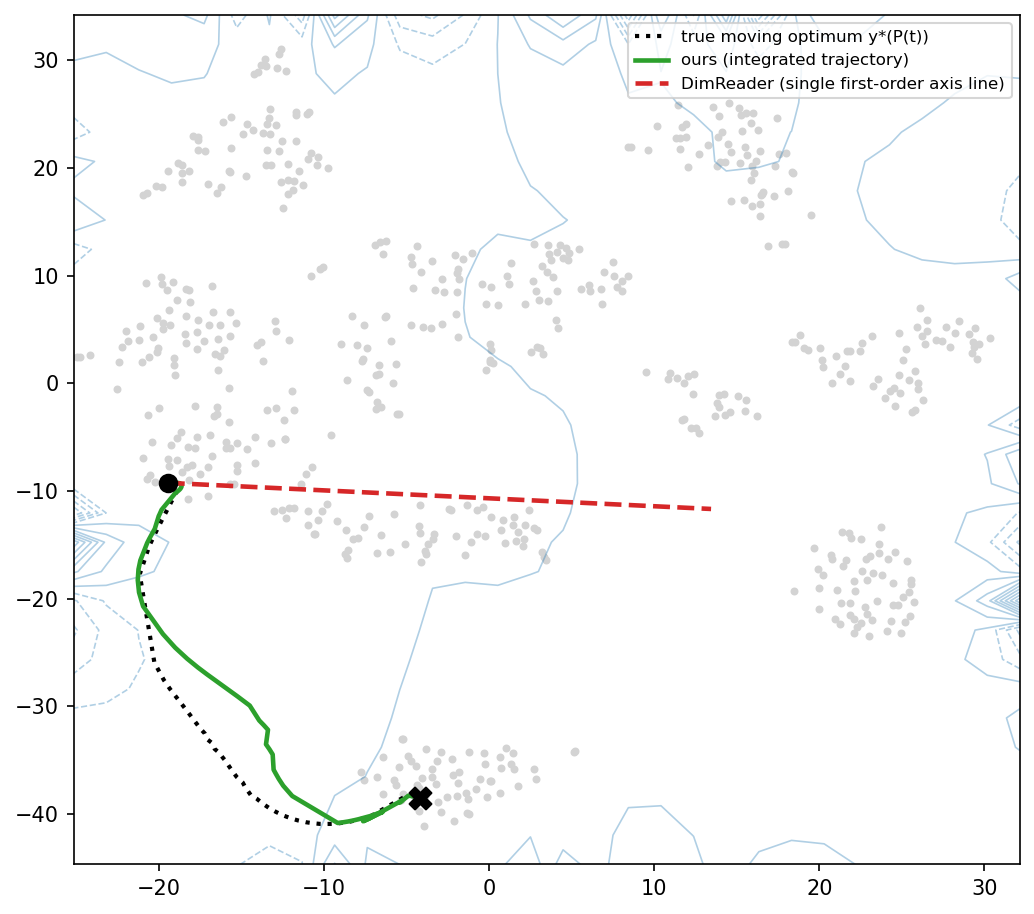}
\caption{Comparison with DimReader~\cite{ref:dimreader}, which approximates the induced
velocity by a single gradient step $v_{\mathrm{DR}}=-\eta\,H_{yx}v$, dropping the
$(H_{yy})^{-1}$ factor.
\emph{Left:} each point is one query on a digits t-SNE map; the horizontal axis is the local
conditioning $\lambda_{\min}^{-1}(H_{yy})$ (rightward $=$ nearer degeneracy; log
scale) and the vertical axis is the angle between DimReader's axis direction and the exact
velocity, with colour encoding $\lambda_{\min}$ itself. The rotation grows as the mode softens
(Pearson $r=0.80$; median error $16^\circ$ in the softest third of queries versus $4^\circ$ in
the stiffest). \emph{Right:} carrying one point along a real inter-cluster path, our integrated
trajectory (green) tracks the true moving optimum $y^\ast(P(t))$ (black dotted) to the target
($\times$), whereas DimReader's single first-order axis line (red), straight by construction,
does not (endpoint error $\approx0$ versus $32$).}
\label{fig:dimreader}
\end{figure}

The exactness is measurable against the axis-based surrogates: dropping the $(H_{yy})^{-1}$ factor
those tools omit visibly rotates the first-order direction near degeneracy, and sends their
integrated axis line off the moving optimum where our trajectory reaches its target
(Fig.~\ref{fig:dimreader}).

\subsection{Level 2: the curvature}\label{sec:curvature}
The connection's \emph{differential} reading differentiates the smooth branch $y^\ast(P)$ in the
well-posed interior. Its first observation is the glyph $J$ above; its second
is the derivative of the first: the induced \emph{map's} second fundamental form, $K=\nabla_P
J$, or in a direction $v$ the directional second derivative $K[v]=d^2y^\ast/dP^2[v,v]$, a vector
in the embedding space.

\smallskip\noindent\emph{Map curvature versus connection curvature.} $K$ curves while $\omega$ stays
flat: it is the second fundamental form of the induced \emph{map} $P\mapsto y^\ast(P)$, not of the
transport, which is curvature-free wherever the branch is single-valued.
Hence $K$ bounds the departure of a finite step from its linearization, while the holonomy
(Sec.~\ref{sec:holonomy}) reports a defect no loop integral of $K$ can recover (Cor.~1,~3).

\smallskip
Throughout, $\|\cdot\|$ on such
embedding-space vectors is the Euclidean ($\ell_2$) norm, and we write the dimensionless relative
curvature in a direction $v$ as $\mathrm{relK}:=\|K[v]\|/\|Jv\|$, the size of the second-order
term against the first-order velocity. Differentiating the optimality condition a second time along the
tangent $(v,Jv)$ yields a clean identity, $K[v]=-H_{yy}^{-1}a$, in which $a$ is the second
directional derivative of $\nabla_y f$ along that tangent, a double Jacobian-vector product
through the argmin. 

\smallskip\noindent\textbf{Proposition P1 (glyph-fidelity certificate).} \emph{On a segment
where $H_{yy}$ stays nonsingular, the first-order glyph prediction $y^\ast+sJv$ departs from the
true position by}
\begin{equation*}\label{eq:P1}
\big\|y^\ast(P+sv)-(y^\ast+sJv)\big\| \;\le\; \tfrac12\,s^2\,\sup\|K[v,v]\| \;+\; O(s^3).
\end{equation*}

\noindent This is the Taylor-remainder form of an $s^2$ law (proof in App.~\ref{app:proofs}) with the $\tfrac12\|K\|$ coefficient remaining correct to order,
empirically verified in Fig.~\ref{fig:curv} (App.~\ref{app:proofs}).
The certificate tells an analyst how far the
cheap glyph reading can be trusted before a full trajectory is needed, certifying the radius of a
placement query.

The map curvature measures the optimization objective, not the data manifold: on two uncontrolled
testbeds its relative value is an order of magnitude larger on the flat plane ($0.597$, MDS) than on
the curved Swiss roll ($0.054$).

\subsection{Counterfactual placement along named analyst axes}\label{sec:diff-scenario}
The everyday question on a single-cell map is \emph{where
would this cell go}. An analyst rarely reads in an arbitrary direction but along a
\emph{named}, interpretable axis --- a marker-contrast direction such as ``as we move from the
B-cell pole toward the CD4 T-cell pole'' --- and what they want from the map is an answer to a
hypothetical: if this cell moved a step $s$ along that axis, where
would this map place it?

The differential reading answers it with two objects, both from the same two derivatives of one
connection. The glyph returns a \emph{location}, $y^\ast(P)+sJv$, from one linear solve and no
optimization; P1 returns the step out to which that location holds, the departure from the true
placement $y^\ast(P+sv)$ being at most $\tfrac12\|K[v]\|s^{2}$. Re-solving the conditional problem
at $P+sv$ is exact but costs one optimization per query, and the certified radius is what says when
that cost is necessary; Sec.~\ref{sec:adv} measures the same trade along a path, where the queries
are consecutive and the saving accumulates.

\smallskip\noindent\emph{What the measurement establishes.} The predicted location is by construction
the first-order term of the true one, so only the radius needs validating: that $\|K[v]\|$ tracks the
actual departure at finite $s$ with P1's $\tfrac12$ coefficient. The ground truth is obtained by
constructing $P+sv$ and re-solving the conditional problem.

\begin{figure}[t]\centering
\includegraphics[width=\columnwidth]{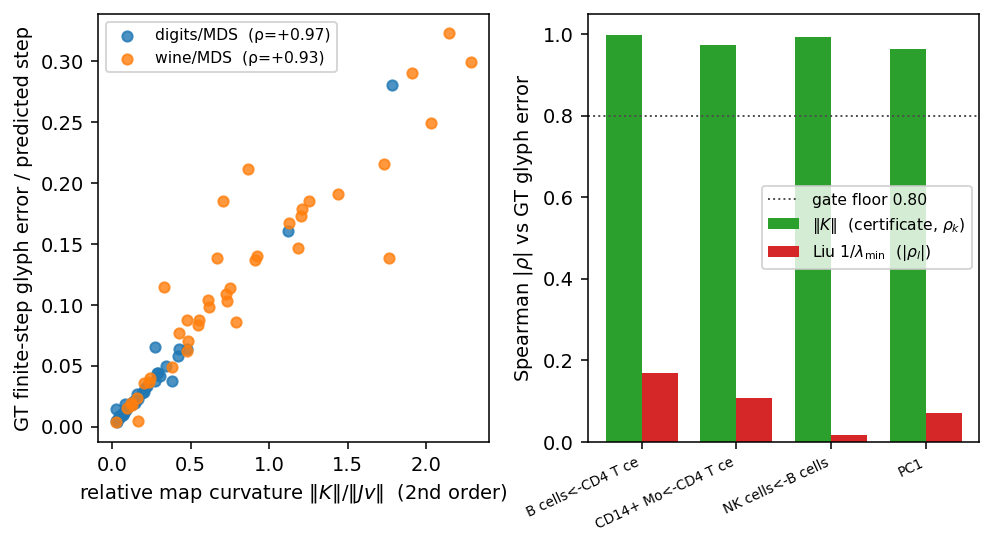}
\caption{The second-order curvature certifies the radius of a counterfactual placement query.
\emph{Left:} on two real MDS embeddings (digits and standardized wine), the second-order $\|K\|$
predicts the ground-truth finite-step departure from the predicted placement (tight diagonal),
while Liu et al.'s direction-blind point-level $\lambda_{\min}^{-1}$ does not.
\emph{Right:} on PBMC3k ($50$-d PCA, MDS) along \emph{named} marker-contrast axes, $\|K\|$ tracks
that departure on every axis (green, above the dotted $\rho\!=\!0.7$ reference) while the
point-level score does not (red). Protocol and per-axis bootstrap CIs in
Sec.~\ref{sec:diff-scenario}.}
\label{fig:curvsowhat}
\end{figure}

We first isolate the claim on two clean MDS embeddings where the direction is the one toward the
query's nearest neighbour, handwritten digits and standardized wine
measurements~\cite{ref:uci}: the second-order $\|K\|$ predicts the ground-truth finite-step glyph
error almost exactly (Spearman $\rho=0.97$ and $0.93$, Fig.~\ref{fig:curvsowhat}, left). The map is
uniformly reliable at the point level, yet the radius varies by an order of magnitude across
directions, and a direction-blind conditioning score is by construction insensitive to that
variation (here Liu et al.'s $\lambda_{\min}^{-1}$ gives $\rho=-0.02$ and $-0.27$): the Level-0
gate of Sec.~\ref{sec:level0} and the Level-2 reading answer different questions.

The query runs along biologically native directions
(Fig.~\ref{fig:curvsowhat}, right). On PBMC3k in its $50$-dimensional PCA representation (the 
common reprocessing step), we take $v$ to be cell-type-contrast axes, centroid differences in PCA space
(B\,$\leftarrow$\,CD4\,T, CD14\,Mono\,$\leftarrow$\,CD4\,T, NK\,$\leftarrow$\,B), a proper
marker-like axis of biological variation, plus the dominant PC. The working set is an independent
stratified-random draw of $141$ cells (all eight types, the rarest capped by its own size); each of
the four axes is read leave-one-out at $35$ query cells. The second-order $\|K[v]\|$ tracks the
measured glyph error along every named axis, with Spearman $\rho$ from $0.963$ to $0.999$ (per-axis
bootstrap $95\%$ CIs, in the order the axes are named above, $[0.992,1.000]$, $[0.923,0.990]$,
$[0.970,0.998]$, $[0.894,0.987]$), while a direction-blind point-level score does not track this
per-direction variation ($|\rho|\le0.19$, every interval spanning zero); the certified step itself
varies by an order of magnitude across axes.

\section{The Integral Reading}\label{sec:integral}
\subsection{Open-path transport: the induced flow}\label{sec:flow-primitive}
The induced connection is also a visual object, and it yields a visualization primitive the
per-point glyph cannot. Where the glyph renders the \emph{static},
per-point transformation of a local subspace, integrating the induced velocity $Jv$ over a region
gives a continuous \emph{flow field} over the embedding, which is the integral counterpart of the glyph and an object a single point cannot express.

\begin{figure}[t]\centering
\includegraphics[width=0.45\columnwidth]{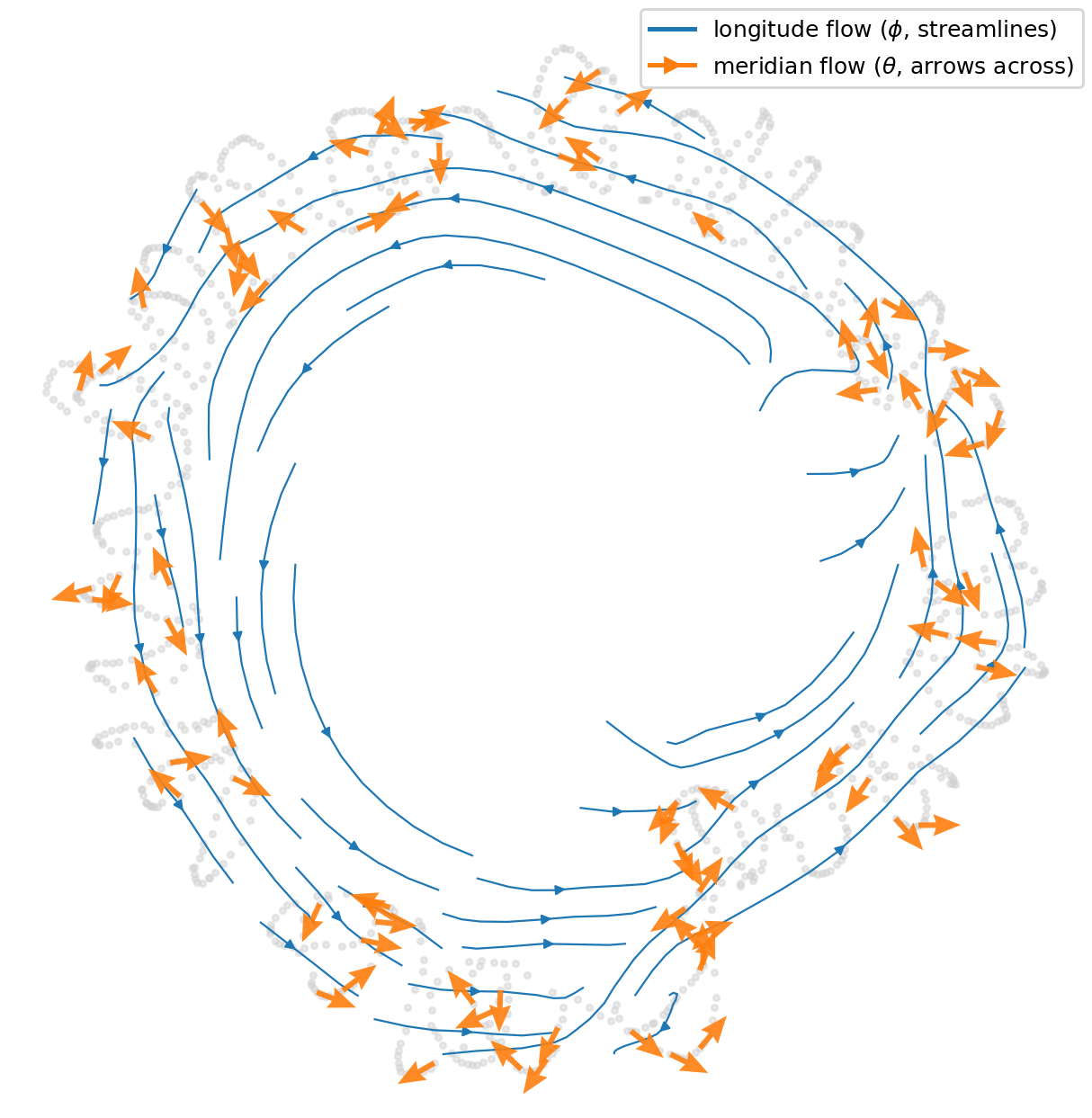}
\caption{The induced flow field as a visualization primitive. A torus embedded by UMAP unrolls to an
annulus, preserving the longitude cycle as the ring and cutting the meridian. The induced longitude
flow $Jv_{\phi}$ circulates around the ring (blue streamlines); the induced meridian flow
$Jv_{\theta}$ points across it (orange arrows), still crossing the ring the cut severed. The two read
as roughly orthogonal. Both channels are masked to the data (the empty hole carries none).}
\label{fig:torusflow}
\end{figure}

Its first use is to read structure a DR cut has hidden (Fig.~\ref{fig:torusflow}). A genus-one torus
cannot be flattened by UMAP without a cut: the embedding \emph{severs} one of its two intrinsic
cycles and lays the surface out as an annulus whose inner and outer boundaries are one and the same
circle on the torus --- an identification the picture gives an analyst no way to see. Reading
\emph{both} cycles through the induced flow puts it back --- the longitude and meridian flows read as
roughly orthogonal, the severed cycle's flow still crossing the ring the embedded geometry no longer
connects. The torus's two independent cycles are thereby recovered as directions on the map the
analyst is already reading: nothing is re-embedded, no second layout has to be reconciled with the
first, and the overlay comes from the same connection that produces the glyph.

The torus cut is \emph{benign}: $H_{yy}$ stays nonsingular and the flow simply transports through it.
Where a tracked path instead crosses the caustic $\Sigma=\{\det H_{yy}=0\}$, $Jv=-H_{yy}^{-1}H_{yx}v$ diverges as $\lambda_{\min}\!\to\!0$
and the objective-consistent placement \emph{tears} --- the hero of
Fig.~\ref{fig:tear}, where $\lambda_{\min}$ collapses by orders of magnitude and $\|Jv\|$ spikes as
the morph crosses the empty region between two clusters. The tear is
a property of the optimization construction. Across $54$ cases --- five well-separated class pairs plus one overlapping control, each under
t-SNE, UMAP, and metric MDS, at three seeds --- it fires on all $30$ t-SNE and UMAP positives
($\lambda_{\min}<0.1$; largest observed $0.058$) and never on metric MDS, whose $\lambda_{\min}$ stays
above $9.67$: a $168\times$ separation between constructions.\footnote{Because $\lambda_{\min}$
carries the scale of the objective that defines it, the invariant content is the within-method
collapse against the same embedding's own well-conditioned baseline; the cross-method contrast is
read at each method's standard loss and reported as a qualitative separation.} The overlapping
control fires as well, confirming that the caustic belongs to these two objectives rather than to
any particular class pair; the specificity is in the \emph{method}.

\smallskip
The open-path reading locates rather than scores: it transports through a benign cut and marks a
singular one, but the magnitude is left ill-determined by the near-singular inverse, and a per-point
score would not survive step refinement. Coverage --- whether a drawn location is backed by data ---
is a separate question, answered by the density check of Sec.~\ref{sec:inverse}.

\subsection{Closed-loop holonomy and the integrability dichotomy}\label{sec:holonomy}
The connection's \emph{integral} reading accumulates transport along finite paths and asks: does
carrying a point around a closed high-d loop return its embedding image to the start? The
flow is the numerical realization of $\omega$, and the loop-closure gap of that integration is the
holonomy. This shifts the question from \emph{point reliability} (the
local faithfulness of a point's neighborhood) to \emph{process reliability}: a faithful embedding
should represent \emph{states}, not \emph{histories}. The neighborhood-preservation
and local-conditioning diagnostics an analyst reaches for answer the former by construction; the
integral reading is the certificate for the latter. We center the \emph{closed}-loop case because it
alone carries a coordinate-free, reparametrization-invariant invariant (Thm.~1(iii)):
a semantic cycle, such as an object's $0^\circ\!\to\!360^\circ$ rotation, should transport back to
itself.

The integral reading has a certifiable direction. Interior transport is flat, so on a disk where the
conditional objective is strongly convex the holonomy is \emph{identically zero}: a
path-independence certificate (P2\,(i)). The non-zero case is where the
reading matters. It makes operational an otherwise global, hard-to-inspect fact that the
conditional objective is non-convex somewhere the loop probes, turning ``the objective is
non-convex'' (a statement about $f$) into ``this OOS placement is path-dependent'' (a statement about
the point an analyst is reading). Formally the non-zero side is a branch obstruction: the transport
is exact on each branch, and what fails is the tracked minimizer's extension across the caustic
$\Sigma=\{\det H_{yy}=0\}$ (App.~\ref{app:proofs}). We call that loop-closure gap the \emph{holonomy}
by analogy (Cor.~3).

The raw loop gap is discretization-dependent, so we read the \emph{branch-distance}: snap the
integrated endpoint to the nearest local minimum of $f(P_0,\cdot)$ and measure its distance to the
starting minimum. Under step refinement this converges (to $\approx0.15$ for t-SNE, to numerically
zero for metric MDS), separating \emph{exact} (well-posed) from \emph{non-exact} (ill-posed) induced
transport (Fig.~\ref{fig:holo}); the verdict is invariant to every protocol knob we sweep while the
raw gap is not (App.~\ref{app:protocol}). Step-refinement convergence is then a gate licensing three
verdicts: \emph{consistent} (holonomy $\equiv0$), \emph{obstructed} (a step-stable branch switch
across $\Sigma$), or \emph{uncertified}, the last when the loop grazes the caustic and collapses
$\lambda_{\min}(H_{yy})$ or the competing-basin gap --- the same gate that disciplines the real COIL
loop (App.~\ref{app:coil}) into a non-closure claim.

\smallskip\noindent\textbf{Proposition P2 (integrability dichotomy).} \emph{Let
$\omega=J\,dP=-H_{yy}^{-1}H_{yx}\,dP$ be the induced-transport one-form and $\Sigma\subset P$ the
caustic over which the tracked minimizer degenerates. Wherever a nondegenerate
minimum branch $y^\ast(P)$ is defined, $\omega=dy^\ast$ is exact, so $\oint_\gamma\omega=0$ for
every loop $\gamma$ bounding a disk on which that branch persists. Consequently: \emph{(i)} if
$f(P,\cdot)$ is strongly convex in $y$ throughout a filling disk of $\gamma$ (unique minimum,
$H_{yy}\succ0$), the branch-distance is identically zero; \emph{(ii)} a non-zero branch-distance
certifies that the loop cannot bound such a disk: the induced transport is not single-valued
over the enclosed region, because the flow met $\Sigma$ or the
terminal snap crossed into a competing basin. The value is a continuous, loop-dependent magnitude
that grades the degree of non-integrability (proof and scope in App.~\ref{app:proofs}).}

\smallskip
The phenomenon is cleanly
cross-method separable on controlled synthetics (t-SNE non-integrable, MDS integrable to
within $4\times10^{-3}$), so an OOS placement is path-dependent under t-SNE and path-independent
under MDS. The non-zero side is existential over loops: it certifies
that \emph{this} loop fails to close --- the query an analyst actually holds --- and other basepoints of the same embedding can read integrable
(App.~\ref{app:protocol}).

\begin{figure}[t]
\centering
\includegraphics[width=\columnwidth]{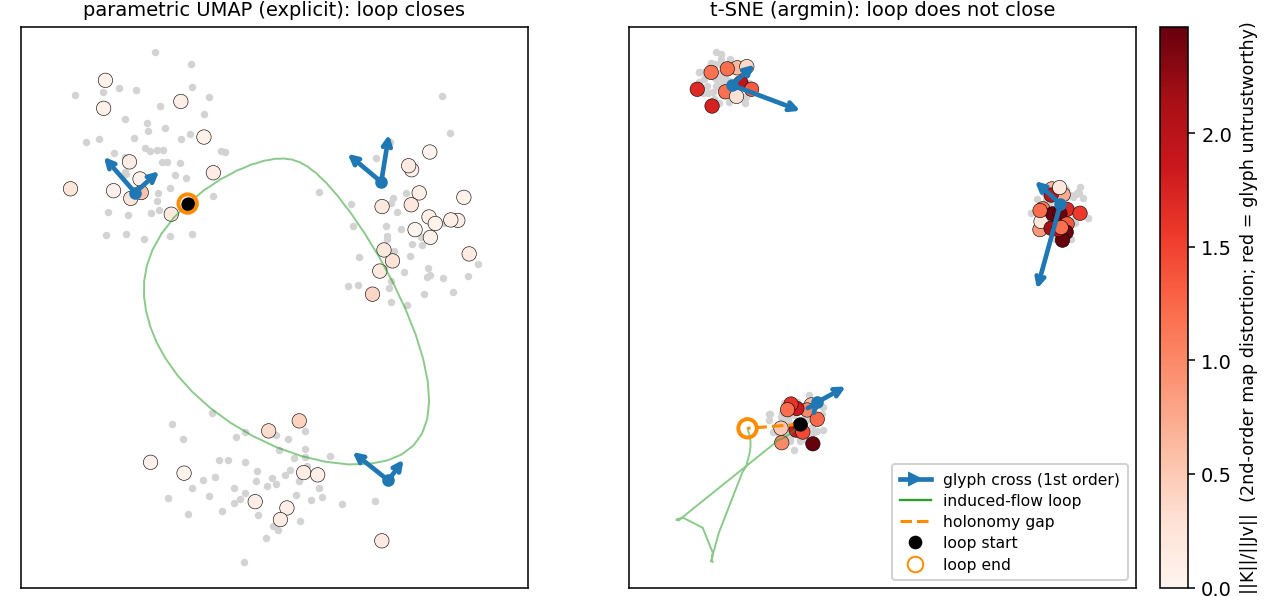}
\caption{\textbf{The integrability dichotomy: both readings, both constructions, one loop.} Two
embeddings of the same data, read around the same closed high-d loop: an \emph{explicit}
parametric UMAP map (left) and an \emph{optimization} t-SNE embedding (right).
\emph{Differential reading (both panels):} a blue \emph{glyph cross} (the induced low-d
velocity of two canonical high-d directions) at the induced position, surrounding
points coloured by the relative map curvature (red $=$ the glyph's
linear reading is untrustworthy) --- alive on \emph{both} constructions.
\emph{Integral reading (the dichotomy):} the green \emph{induced-flow loop}
transports one point around the loop; the explicit map returns to its start, the t-SNE
argmin flow does not, the dashed orange gap measuring the non-closure. Gray points: the remaining
embedding.}
\label{fig:dichotomy}
\end{figure}

\smallskip\noindent\emph{Single-valued maps versus branch-obstructed argmins.}\label{sec:dichotomy}
The connection is defined for any differentiable embedding, but the two constructions of
Sec.~\ref{sec:operator} sit on opposite sides of the integrability dichotomy, and the difference is
measurable. On the same data and the same closed high-d loop we contrast an \emph{explicit}
trained map $\Phi$ (\emph{parametric UMAP}) against an \emph{optimization} embedding (t-SNE),
integrating with the identical Runge--Kutta scheme (Fig.~\ref{fig:dichotomy}). The explicit map
returns its loop image, holonomy $=0$ --- structurally, since $\omega=d\Phi$ is exact and has no
caustic on any loop --- yet its curvature is non-zero ($\mathrm{relK}>0$): the differential reading
stays informative while the integral one vanishes. The optimization embedding does not close the
loop. That the vanishing tracks \emph{single-valuedness} rather than the objective family is
confirmed by a second, structurally unrelated explicit family, a 2D-bottleneck \emph{autoencoder},
returning holonomy $=0$ as well (Table~\ref{tab:ladder}). The reading therefore tests whether the
induced argmin is single-valued: a non-zero value \emph{certifies} a non-convex argmin with a caustic
the loop encloses --- no explicit map can produce one --- whereas zero marks a single-valued induced
map, which an explicit map and a convex objective (MDS) both realize; among optimization objectives
the value is itself graded by non-convexity (zero for MDS, non-zero for t-SNE, Cor.~1). The explicit
side doubles as a control: the identical integrator yields zero on a genuinely single-valued
map, so the non-zero holonomy read elsewhere is a branch obstruction.

The dichotomy holds on real data too: on COIL-20's object-rotation cycle\, a
held-out autoencoder closes the loop while the t-SNE argmin flow does not (App.~\ref{app:coil}).

\subsection{What the certificates see: the conditional loss landscape}\label{sec:landscape}
\begin{figure*}[t]\centering
\subfloat[Static: one fixed ambiguous query.]{%
\includegraphics[width=0.4\textwidth]{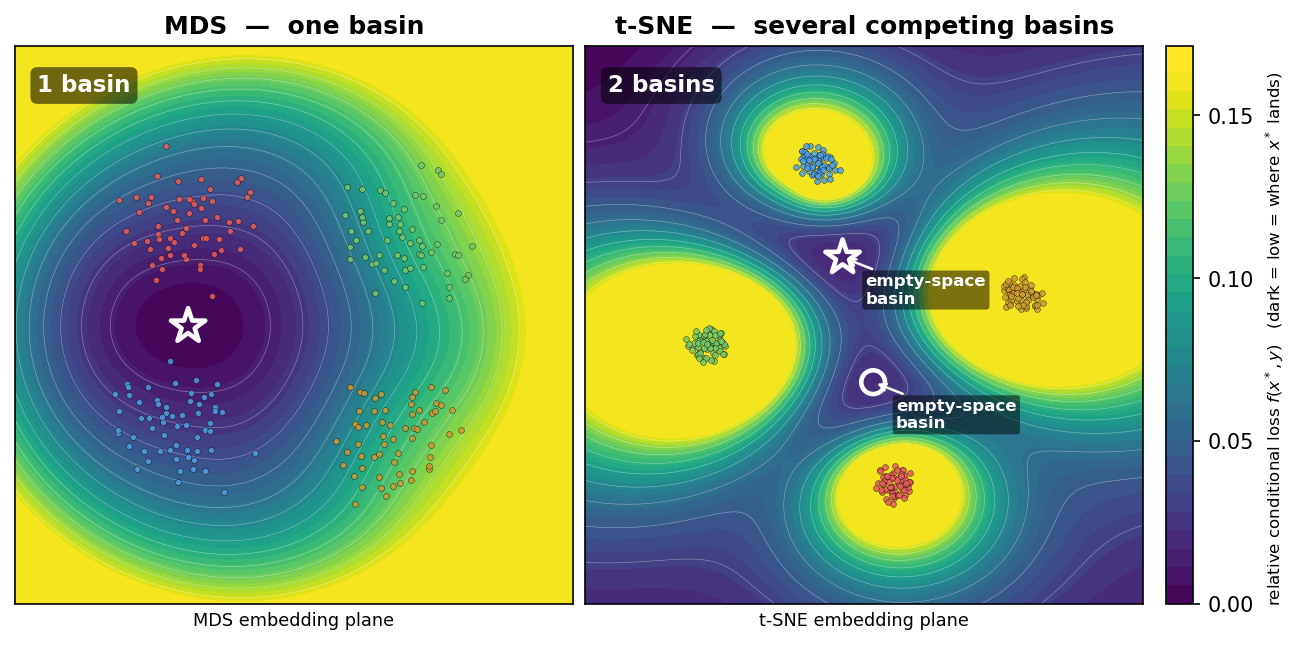}}\hfil
\subfloat[Dynamic: sweeping that query, cluster A to B.]{%
\includegraphics[width=0.43\textwidth]{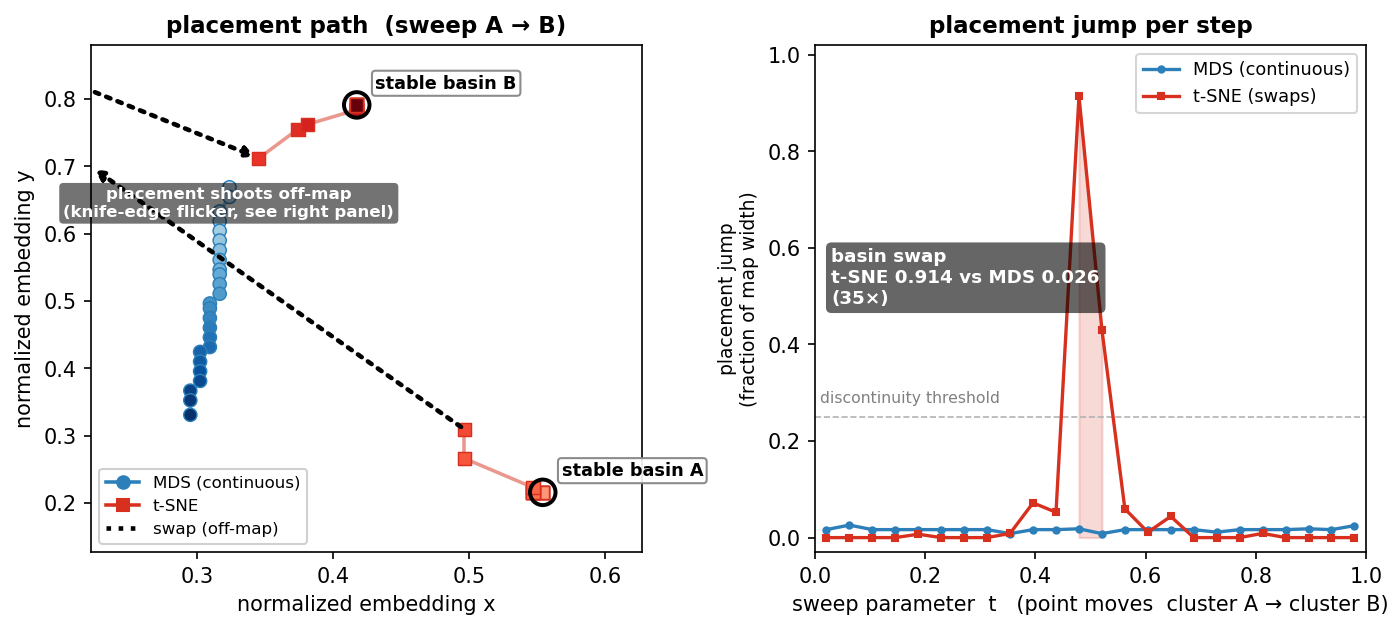}}
\caption{The conditional loss landscape $f(x^\ast,\cdot)$ --- the objective the induced reading
differentiates --- drawn as a map of the embedding plane. \emph{(a)} Color is the relative
conditional loss; the star marks the global optimum,
open circles the competing basins, small dots the
anchor embeddings colored by cluster. MDS is a single bowl; t-SNE is multi-basin. \emph{(b)} As
$x^\ast$ sweeps along a high-d path, the
t-SNE placement (red) jumps between stable basins while MDS (blue) glides; the per-step jump (right)
crosses the plotted discontinuity threshold of $0.25$ (fraction of map width) only for
t-SNE ($0.914$ vs.\ MDS $0.026$).}
\label{fig:landscape}
\end{figure*}

The operative regime can be drawn. The object is the conditional loss
landscape $f(x^\ast,\cdot)$ the induced reading differentiates (Fig.~\ref{fig:landscape}), and drawing
what the framework already computes makes a structural point: \emph{Liu et al.'s $\lambda_{\min}^{-1}$
is the curvature of one basin}, whereas what decides an OOS placement is which basin among several the
point falls into.

Read statically, for one fixed ambiguous query the landscape is a single bowl for MDS but multi-basin
for t-SNE, whose competing minima include \emph{empty-space} basins between clusters that no data
occupies yet the objective rewards. Read dynamically, sweeping the query along a high-d
path, the MDS placement glides while the t-SNE placement \emph{jumps} between stable basins. Those
jumps are the basin-swaps of $\{\text{argmin non-unique}\}$ that P3 places outside the differential
reading's reach and the holonomy inside it: the landscape is the visual form of
\emph{local}\,$\subset$\,\emph{global}, the boundary at which our reading and Liu et al.'s coincide
locally but diverge globally.

\section{The Rank-Based Metrics at Finite Scale}\label{sec:finitescale}
The differential reading is pointwise, yet the quality metrics an analyst reaches for, \eg trustworthiness,
continuity, co-ranking, kNN recall, are finite-sample, finite-scale statistics of neighbour rankings.
They are \emph{not} functions of the local jet: two embeddings sharing every local derivative at a point
can still differ in trustworthiness through the placement of \emph{other} samples (a bump supported away
from the point moves a far neighbour without touching the jet; App.~\ref{app:finitescale} makes this
irreducibility precise). We call these classical metrics the \emph{relational} reading and show that although the differential reading
cannot itself report it, it \emph{controls} the finite-radius distortion that governs when a ranking can flip.

\smallskip\noindent\textbf{Lemma (finite-scale distortion).} \emph{Let $\Phi\in C^2$, $J_i=d\Phi$, and
let $\alpha_iQ_i$ be the nearest scaled isometry to $J_i$ on the data-manifold tangent, with anisotropy
$\delta_i=\|J_i-\alpha_iQ_i\|$ (the glyph's departure from a similarity) and curvature bound
$M_i(R)=\sup_{B_R}\|K\|$, the same second fundamental form $K=\nabla_PJ$ of Sec.~\ref{sec:curvature} whose
$\tfrac12\|K\|$ coefficient controls the glyph-fidelity certificate~P1. Then for $\|x_j-x_i\|\le R$,
$\;\bigl|\hat d_{ij}-\alpha_i d_{ij}\bigr|\le \epsilon_i(R)=\delta_iR+\tfrac12 M_i(R)R^2.$}

\smallskip\noindent\textbf{Proposition~P5 (finite-scale neighbourhood preservation).} \emph{With
$d_{i,(k)}$ the $k$-th high-d neighbour distance and margin
$\Delta_{i,k}=d_{i,(k+1)}-d_{i,(k)}$, if every candidate that could enter or leave the $k$-neighbourhood
lies within $R$ and $\alpha_i\Delta_{i,k}>2\epsilon_i(R)$, then $N_k^X(i)=N_k^Y(i)$, hence
$T_k(i)=C_k(i)=1$. Moreover $N_k^X(i)\,\triangle\,N_k^Y(i)\subseteq\{j:|d_{ij}-d_{i,(k)}|\le
2\epsilon_i(R)/\alpha_i\}$: every rank error is confined to an ambiguity band, bounding
trustworthiness/continuity loss (proof in App.~\ref{app:finitescale}).}

\smallskip
Finite-neighbourhood reliability thus decomposes as glyph anisotropy ($\delta$) $+$ map curvature
($M\!=\!K$) $+$ observation scale ($R,k$) $+$ sample margin ($\Delta$): the two differential invariants
this paper already reads are the geometric inputs, while the finite-sample piece the differential
reading provably \emph{cannot} supply is the margin $\Delta$, the classical metrics' irreducible content.

On a controlled analytic testbed (a near-identity map with a localized anisotropic bump, $n=1500$,
$k=10$) the condition is both non-vacuous and sound where it matters: it certifies $1208$ of $1500$
points, and of the $36$ points whose embedded $k$-neighbourhood genuinely differs from its
high-d one it certifies \emph{none}. The certified fraction reflects this testbed's
mildness. P5 certifies where preservation is provably exact.

\section{The Instrument and Its Use}\label{sec:instrument}
\subsection{The linear limit and an ordering over methods}\label{sec:generator}\label{sec:ladder}
The framework's quantities are two readings indexed three ways --- by order, by scale, and by
direction. The glyph and the curvature are the first and second order of one differential reading,
where the order stops at two by the reach of Theorem~1; open-path and closed-loop transport are one
integral reading at two scales; and the direction is an axis across both readings, the same $J$
read through $J_T^{+}$ transporting backward wherever it transports forward
(Sec.~\ref{sec:inverse}, Table~\ref{tab:bidirectional}).

The same structure orders the \emph{methods} an analyst might use, and it reaches outside nonlinear
DR. Read principal component analysis as a conditional
reconstruction objective $f(x,y)=\|x-Wy\|^2$ with $W$ the fixed orthonormal projection basis;
then its representation is $J=dy^\ast/dx=W^\top$, exactly the projection matrix, so the
glyph specializes to the linear map on linear DR. Because that objective is \emph{jointly}
quadratic in $x$ and $y$, both
higher readings vanish \emph{identically}: the second fundamental form $K=\nabla_P J$ is
machine-zero and the holonomy is machine-zero (Table~\ref{tab:ladder}); the connection is flat,
and in fact constant. Linear DR is the flat corner of the framework which is directly useful: projecting 
a high-d trajectory into a low-d view through a fixed projection matrix, as recent GapMiner
does~\cite{ref:void}, is the integral of this first-order reading along the path.

Holding the data fixed and varying only the objective exposes a clean ladder: for a linear objective
both new readings are machine-zero; for a strongly convex non-quadratic one the curvature can be
non-zero while the unique minimum forces holonomy $\equiv0$; only a non-convex objective admits both
(Table~\ref{tab:ladder}, proof in App.~\ref{app:proofs}). The set of active readings therefore grows
monotonically with the objective's structural complexity, bounding the readings a given method
obliges an analyst to consult before the point-level regimes below apply.

\begin{table}[t]\centering
\caption{Reading ladder: within the optimization family each reading switches on with
objective complexity. The last two rows swap the \emph{construction} to explicit maps.
Protocol, identical across rows: the same data and query
point; the curvature column is $\max\|K[v]\|/\|Jv\|$ over twelve random unit directions $v$, the
integral column the loop gap normalized by the embedding diameter, so both are dimensionless. 
}
\label{tab:ladder}
\begin{tabular}{lccc}
\toprule
& \multicolumn{2}{c}{differential} & integral\\
\cmidrule(lr){2-3}\cmidrule(lr){4-4}
method & glyph $J$ & curvature $\mathrm{relK}$ & holonomy/diam\\
\midrule
PCA / linear   & $W^\top$ (const) & \textbf{0} & \textbf{0}\\
metric MDS     & varies & 4.66 & $\approx0$\\
t-SNE          & varies & 4.82 & 0.15\\
\midrule
parametric UMAP & varies & 0.38 & $\approx0$\\
autoencoder     & varies & 1.49 & $\approx0$\\
\bottomrule
\end{tabular}
\end{table}

\subsection{The four regimes}\label{sec:regimes}
Read together, the two readings expose how a single embedding behaves (Fig.~\ref{fig:dichotomy},
right). The framework answers an analyst's \emph{can I trust where the map places this} with two
certificates of the same $\omega$: the second-order curvature bounds the glyph's finite-step
extrapolation error, and the holonomy reports whether a
placement is route-independent. They pass and fail independently, the integral one structurally
(zero or not, P2\,(i)) and the differential one against the step an analyst cares about, through
P1's $\tfrac12\|K[v]\|s^{2}$. Four regimes are therefore available --- \textbf{(A)} both pass and the
cheap glyph reading is
trustworthy; \textbf{(B)} the differential passes while the integral fails, locally faithful but
globally vague, which is the operative case since nothing local signals it; \textbf{(C)} the
integral passes while the differential fails, so the reading must be integrated rather than
linearized; and the doubly failing corner, where no reading is trustworthy. The two certificates are
asked of different objects.
Therefore, the regimes classify a \emph{reading}.

\subsection{Reading the connection backward}\label{sec:inverse}
Nothing in the account so far was built to answer a backward question, yet it answers one with no new
object and no certificate the framework does not already compute --- evidence that the connection is
the productive unit.

The same connection read forward also transports \emph{backward}. An analyst who draws a low-d
path $\gamma$ on the map and asks \emph{which high-d states does this trace} is asking to invert
$\omega$: to recover a high-d trajectory whose induced image is $\gamma$. Since
$J=-H_{yy}^{-1}H_{yx}$ carries a high-d perturbation to a low-d velocity ($d\times D$,
$d\ll D$), the raw inverse is under-determined; the data manifold selects the answer. Restricting $J$ to
the data-manifold tangent $T_PM$ and inverting on that subspace back-projects $\gamma$ to an on-manifold
trajectory, integrated with the same coupled flow and a manifold corrector.

\smallskip\noindent\textbf{Proposition~P4 (inverse back-transport).} \emph{Let $J_T=J|_{T_PM}$ be the
connection restricted to the data-manifold tangent. Where $J_T$ is full rank, its pseudo-inverse
$J_T^{+}$ back-projects a low-d direction to the on-manifold high-d direction
realizing it; integrating $J_T^{+}\dot\gamma$ recovers the on-manifold trajectory whose induced image is
$\gamma$, exact on a well-posed embedding (where $\omega=dy^\ast$ is flat) and to first order in the step
otherwise.}

\smallskip
On a plane under MDS the recovery is exact --- $J_T=I$ and the back-projected path leaves the plane by 
zero, the flat corner of the reading ladder (Sec.~\ref{sec:ladder}) read backward.

A drawn location fails to have a unique on-manifold preimage in exactly two ways, and the framework
already computes a certificate for each. It
may lie in an \emph{empty} region the manifold does not cover, flagged by a trivial data-density check; or
it may lie over a \emph{fold}, where several high-d states share one placement ---
\emph{precisely} where the induced transport is non-single-valued, \ie where the forward holonomy
is non-zero (P2). Neither is visible to the alternatives: the corrector
keeps the trajectory on the manifold, where $\lambda_{\min}(H_{yy})$ and the glyph conditioning stay
healthy, and a round-trip residual $\|\Phi(x_{\mathrm{rec}})-\gamma\|$ extrapolates over an empty hole
(the objective is anchor-defined there) and returns one valid branch at a fold. On a controlled
fold-and-hole testbed the decomposition is clean (App.~\ref{app:inverse}): density detects the
empty region and the forward holonomy the fold (AUC $1.00$ each), while both differential
certificates and the round-trip residual detect neither (round-trip's incremental AUC over density is
$-0.15$, CI $[-0.28,-0.01]$). \emph{Forward holonomy is the inverse's fold-certificate}
(Table~\ref{tab:bidirectional}).

\begin{table}[t]
\caption{One connection, both directions: forward transport ($J$) and inverse back-transport
($J_T^{+}$) read by the same two readings. The two integral cells share one certificate ---
the \emph{forward} holonomy certifies the \emph{inverse} fold (P4).
}
\label{tab:bidirectional}
\footnotesize
\centering
\begin{tabular}{>{\raggedright\arraybackslash}p{1.15cm} | >{\raggedright\arraybackslash}p{3.0cm} >{\raggedright\arraybackslash}p{2.7cm}}
\toprule
\textbf{Direction} & \textbf{Differential} & \textbf{Integral} \\
\midrule
\textbf{Forward}\newline($J$)          & $\mathrm{relK}$: move read consistently (P1), gated by $\lambda_{\min}$ (Sec.~\ref{sec:level0}) & holonomy: placement route-independent (P2) \\
\textbf{Inverse}\newline($J_T^{+}$)    & $\sigma_{\min}(J_T)$ / round-trip: drawn direction realizable on-manifold & fold $=$ forward holonomy: drawn point has a unique preimage (P4) \\
\bottomrule
\end{tabular}
\end{table}

The two differential cells are distinct loci of the same connection: forward ill-posedness is
$H_{yy}$ near-singular ($\Sigma$, where the velocity blows up), inverse ill-posedness is $J_T$
rank-deficient (a drawn direction with no on-manifold realization). Neither implies the other, so the
inverse needs its own differential cell rather than a re-reading of the forward one.

\smallskip
The back-projection is native to \emph{optimization-based} DR, where no decoder exists and the
connection supplies the inverse for free from the $H_{yy},H_{yx}$ already computed; where a decoder
does exist it inverts the drawn path directly. What the connection contributes is the two
certificates rather than reconstruction quality, and we make no performance claim against the
inverse-projection literature --- local affine schemes~\cite{ref:ilamp}, learned inverse
networks~\cite{ref:nninv}, and applications such as classifier decision-boundary
maps~\cite{ref:dbm}.

\subsection{Use case: tracing a perturbation path through a fixed map}\label{sec:adv}
\begin{figure}[t]\centering
\includegraphics[width=\columnwidth]{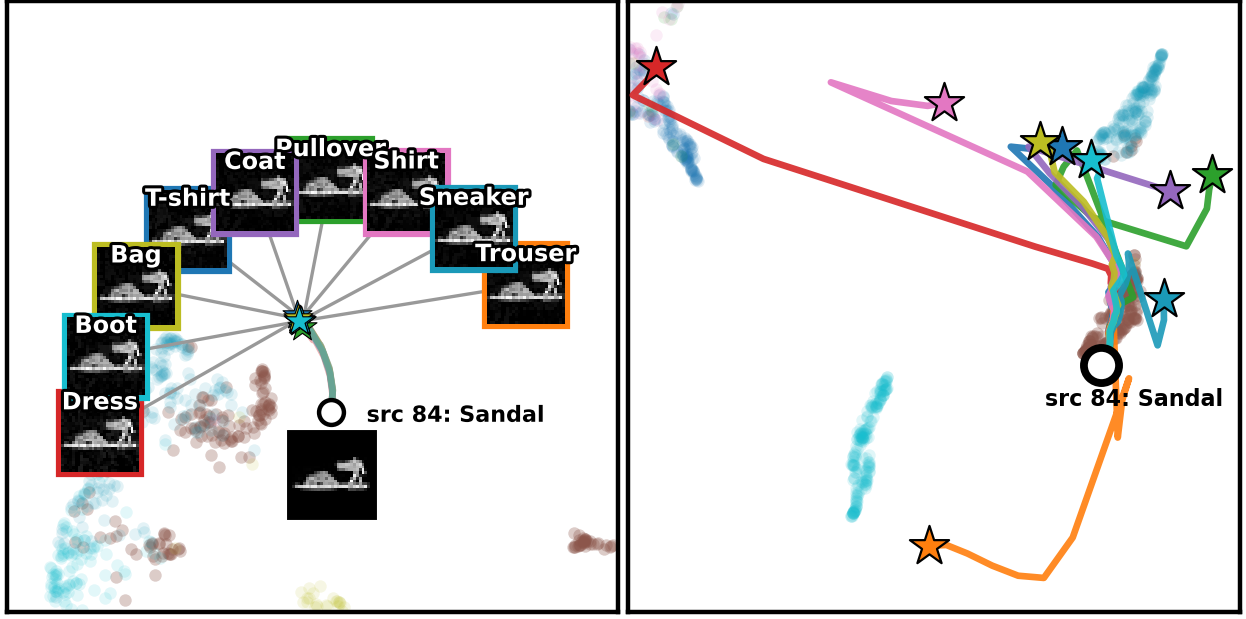}
\begin{minipage}[b]{0.5\columnwidth}\centering
\includegraphics[width=0.80\linewidth]{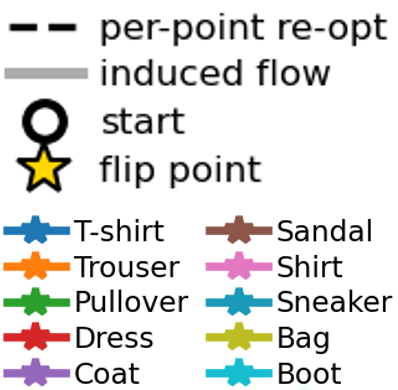}
\end{minipage}%
\begin{minipage}[b]{0.5\columnwidth}\raggedright
\includegraphics[width=\linewidth]{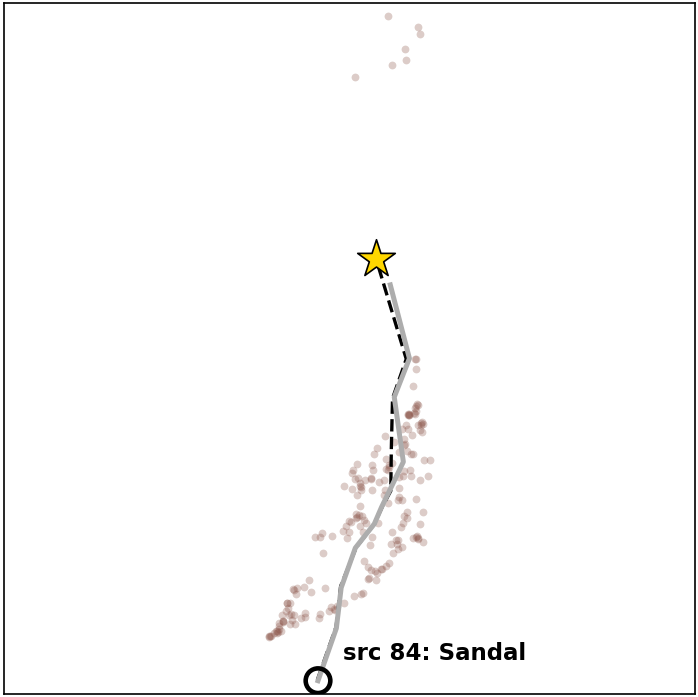}%
\end{minipage}
\caption{\textbf{Reading one adversarial neighbourhood through the induced flow.} A Sandal (src~84) is
attacked toward each of the nine other classes (targeted PGD, all nine flip); the induced flow traces
each morph on a fixed UMAP of the same data.
\emph{Top left (raw image space):} the nine attacks leave through a single seam
($1^\circ$ of angular spread); thumbnails are fanned out for legibility, their borders and labels
giving the class the CNN now reads. Petals are chord-morphs; only their endpoints are real
adversarial images. \emph{Top right (CNN feature space):} the \emph{same} attacks
fan out far more widely.
\emph{Bottom right:} the induced flow against the per-point re-optimization, which corners
between adjacent conditional minima near the flip; the flow matches it at $\approx\!3.6\times$ fewer
objective evaluations and matched faithfulness (App.~\ref{app:advpath}).}
\label{fig:advpath}
\end{figure}

The reading an analyst performs most often is not static: they watch a point \emph{move}, \eg a sample
under an adversarial or generative perturbation, a state along a control input, a cell along a
developmental trajectory, and ask what the fixed map says about the transition. Integrating the
induced flow is the instrument for it. We demonstrate on an adversarial attack
where the reading is both useful and easy to get wrong: attacked toward each of the nine other
classes, one source image leaves a raw-pixel UMAP through a single seam but spreads widely
on a CNN feature UMAP of the same data in
Fig.~\ref{fig:advpath}. Two maps disagree wildly about the geometry of one attack, and an analyst has
to decide how far to trust either. The induced flow is the construction that draws these paths,
placing every frame at the map's own objective-consistent minimum. Concretely, we drive a ResNet-18 input along the pixel-space path, read
its penultimate features frame by frame as OOS queries against a fixed UMAP map, and
integrate the induced flow to obtain the trajectory of the moving placement
(Fig.~\ref{fig:advpath}). We show where an embedding represents an
adversarial path faithfully and where it distorts it.

Three properties, each certified elsewhere in the paper and consolidated here on one path, make this a
usable instrument. \emph{(i) Continuity.} The induced trajectory is a single smooth curve where
re-optimizing each frame independently corners between adjacent conditional minima (the
faithfulness--smoothness knee, Sec.~\ref{sec:eval}, item ii; the Oracle's kinks near the flip), so
reading a transition off the flow does not manufacture the frame-to-frame jitter that independent
re-embedding of each frame can. \emph{(ii) Amortization.} It reaches that placement at
$\approx\!3.6\times$ fewer objective evaluations than re-optimizing every frame from scratch, at
matched faithfulness ($\Delta f$ within tolerance, App.~\ref{app:sanity}). 
\emph{(iii) Effort where it is needed.} The connection's
conditioning $\lambda_{\min}(H_{yy})$ tells the flow \emph{where} a correction back to the
conditional optimum is worth its cost, with a watchdog floor so integration error cannot drift
unchecked along a well-conditioned stretch where $\lambda_{\min}$ never fires.

The conditioning that schedules that effort also bounds the reading.
The three sources above are well-conditioned by design; the boundary lives where a path approaches
the caustic ($\lambda_{\min}\!\to\!0$), the regime Fig.~\ref{fig:tear} isolates. There the
conditional optimum ceases to be single-valued --- a
flat basin of low-$\Delta f$ placements opens and correction saturates --- and the placement
\emph{teleports} instead of gliding, which the instrument \emph{marks} rather than resolves. The
certificate is $\lambda_{\min}(H_{yy})$ at the conditional optimum $y^\star(P)$, and the cheap proxy the
flow evaluates \emph{along the path} can under-report a degeneracy that sits at a frame's own optimum
rather than at the points the trajectory visits.

\section{Evaluation}\label{sec:eval}
Each quantitative headline in Table~\ref{tab:eval} is produced by a module asserting an acceptance
gate; qualitative exhibits are marked as such where they appear.

\emph{(i) Ground-truth fidelity.} On synthetic data with analytic ground truth the induced flow
recovers the correct geometry: an isometric plane under MDS reproduces the ground-truth
tangents, and a swiss roll under Isomap tracks the analytic geodesic. The same family anchors the
rate-ladder verification of Theorem~2.

\emph{(ii) Faithful access to the integral reading.} Along an induced trajectory the conditional loss
stays low (faithful) while the
low-d path stays smooth, placing the flow at the Pareto-efficient knee of the
faithfulness--smoothness trade-off --- as faithful as point-wise re-optimization yet markedly
smoother, and far more faithful than the naive interpolation baselines. That a measured holonomy is
geometric rather than integrator drift is established separately, by the two controls of
Sec.~\ref{sec:holonomy}.

\emph{(iii) Case studies.} The two load-bearing studies instantiate the two readings on real data:
the counterfactual-placement study (Sec.~\ref{sec:diff-scenario}) on the differential side, and on
the integral one the integrability dichotomy (Sec.~\ref{sec:dichotomy}) together with the
perturbation-path scenario (Sec.~\ref{sec:adv}) that puts it to work; the COIL-20 rotation
cycle (App.~\ref{app:coil}) adds a qualitative path-inconsistency exhibit.
Robustness of the quantitative vignettes is checked across five seeds.

\begin{table}[t]\centering
\caption{Evaluation summary.}
\label{tab:eval}
\footnotesize
\setlength{\tabcolsep}{4pt}
\begin{tabular}{@{}lll@{}}
\toprule
item & metric & result\\
\midrule
GT fidelity & tangent / geodesic & recovers analytic GT \\
rate ladder & slope orders & $-0.99$ / $-3.00$ \\
flow faithfulness & loss / smoothness & faithful vs.\ interp / re-opt \\
placement radius & $\rho(\|K\|,\text{GT glyph err})$ & $0.963$--$0.999$ \\
integrability & holo., explicit vs.\ argmin & $\approx0$ vs.\ $0.15$ \\
\bottomrule
\end{tabular}
\end{table}


\section{Discussion and Limitations}

The two readings measure complementary properties of one geometry and are independent by
construction (Corollary~1); neither is a universal embedding-quality score.
The minimality result covers readings of \emph{finite order}
sampled at \emph{finitely many} points, so a diagnostic that reads local data continuously along the
whole path falls outside it (App.~\ref{app:minimality}).

The four regimes (Sec.~\ref{sec:regimes}) give the shape of a reading protocol; We leave
its calibration and a controlled user study quantifying its downstream effect on analyst decisions the future work.

The framework requires only that the embedding be differentiable, and reads both constructions of
Sec.~\ref{sec:operator} identically once the connection is in hand. The one structural difference is
the integral reading, which is vacuous on the explicit
side (Sec.~\ref{sec:dichotomy}); foundation-model encoders fall under that same case, so their
holonomy is zero by the same argument, a theorem about exact one-forms. Nondifferentiable or purely stochastic embeddings are out of scope.

A subtler assumption concerns the \emph{conditional} objective. The induced connection reads an OOS
objective in which a single query moves against a fixed anchor set, and specifying it fixes the
kernel bandwidth (perplexity), the anchor subset, and the softmin/membership calibrations of Isomap
and UMAP. The readings are exact derivatives of the objective \emph{as specified} for the
map the analyst is looking at, not invariant across such constructions. The anchor-subset dependence
is probed by the five-seed robustness of Sec.~\ref{sec:eval}, whose conclusions survive resampling,
and P2's certified direction requires only strong convexity, independent of the bandwidth;
invariance to the bandwidth and temperature we do not establish.

\section{Conclusion}\label{sec:conclusion}
We have shown that a static projection glyph and a point-level map-continuity score are two internally connected positions in one structure, determined by the transport connection
$\omega=J\,dP$ that every differentiable embedding induces --- the glyph as its first-order reading,
the point-level score as the well-posedness gate every reading presupposes. Beyond recovering those
two, the framework adds two observables, both exact
via automatic differentiation: the \emph{differential} reading's second-order curvature (the induced
map's second fundamental form), which certifies finite-step glyph fidelity, and the \emph{integral}
reading's holonomy (the flat connection's branch obstruction across the caustic), which measures
path-consistency. The two are geometrically decoupled and functionally independent (Corollary~1), and
one theorem certifies the readings canonical at each level and the integral level necessary
(Theorem~1). A good framework is not one that introduces more quantities but one that makes the
existing quantities inevitable --- complete at each level, and irreducible at the integral one.

\section*{Acknowledgments}
The author thanks Yuwen Long Esq. for her warm support and companionship throughout this work.

\putbib[refs]
\end{bibunit}   
\begin{IEEEbiographynophoto}{Xinyu Zhang}
earned his B.E. at Shandong University, Taishan College and Ph.D. at Stony Brook University. His research interests include multivariate data analysis, and reinforcement learning.
\end{IEEEbiographynophoto}
\begin{IEEEbiographynophoto}{Klaus Mueller}
is a Professor of Computer Science at Stony Brook University and a senior scientist at Brookhaven National Lab. His research interests include explainable AI, visual analytics, data science, and medical imaging. 
To date, his 300+ papers have been cited over 15,500 times. 
He is a IEEE Fellow.
\end{IEEEbiographynophoto}
\input{appendices}
\end{document}

%% file: appendices.tex
\clearpage

\appendices
\begin{bibunit}[IEEEtran]   
\section{Proofs of the Propositions}\label{app:proofs}
Throughout, $f\in C^{3}$ in a neighborhood of $(P_{0},y^{\ast}(P_{0}))$, and on the relevant
set $H_{yy}=\nabla^{2}_{yy}f$ is nonsingular, so the branch $P\mapsto y^{\ast}(P)$ is $C^{2}$ by
the implicit function theorem (IFT). We write $J=-H_{yy}^{-1}H_{yx}$, $K[v]=d^{2}y^{\ast}/dP^{2}[v,v]$,
$\lambda_{\min}=\lambda_{\min}(H_{yy})$, and $D_{\mathrm{loc}}=\{\det H_{yy}=0\}$.

\subsection{Theorem~1 (i)--(iii): canonical representation}\label{app:representation}

\smallskip\noindent\emph{The naturality conditions.} Call a diagnostic \emph{embedding-induced} if it
depends on the embedding only through the induced map $\Phi=y^{\ast}$ (equal $\Phi$ $\Rightarrow$
equal value); \emph{local of order $r$} if its value at $P$ depends only on the $r$-jet
$j^{r}_{P}y^{\ast}$; and \emph{coordinate-equivariant} if a Euclidean isometry of the embedding
coordinates $y\mapsto Qy+b$ ($Q^{\top}Q=I$) acts on it only through the induced action on the
corresponding tensors --- $J\mapsto QJ$, $K\mapsto QK$, so an $\ell_{2}$ scalar such as
$\mathrm{relK}$ is invariant, and a translation acts trivially on local quantities. A path diagnostic
is \emph{path-local and reparametrization-invariant} if it depends on $y^{\ast}$ along $\gamma$ only
through the transport $\dot y=J\dot\gamma$ and is unchanged by orientation-preserving
reparametrization. ``Natural'' in Sec.~\ref{sec:representation} means exactly this list.

\smallskip
\emph{(i) First order.} An embedding-induced diagnostic depends on $f$ only through $y^{\ast}$; being
local of order $\le1$ its value at $P$ depends only on
$j^{1}_{P}y^{\ast}=(y^{\ast}(P),d_{P}y^{\ast})$, so $D_{1}(P)=\Phi_{1}(y^{\ast}(P),d_{P}y^{\ast})$.
Coordinate equivariance under translations $y\mapsto y+b$ forbids dependence on the absolute
position $y^{\ast}(P)$, leaving $D_{1}(P)=\Psi_{1}(d_{P}y^{\ast})=\Psi_{1}(J)$ with
$J=-H_{yy}^{-1}H_{yx}$ by the IFT. Being scalar, $D_{1}$ is moreover \emph{invariant} under the
rotation action $J\mapsto QJ$, not merely equivariant, so it factors through the complete
$O(d)$-invariant of the columns of $J$ --- the pullback metric $J^{\top}J$ (two glyphs share it iff
they agree up to some $Q\in O(d)$). The coordinate glyph $J$ is the equivariant representative;
$J^{\top}J$ is the frame-free content a scalar reading actually sees (directional stretch, singular
values, anisotropy, local volume distortion).
\emph{(ii) Second order.} Order-$\le2$ locality gives
$D_{2}=\Phi_{2}(y^{\ast},d_{P}y^{\ast},d^{2}_{P}y^{\ast})$; equivariance removes $y^{\ast}$, and
$d_{P}y^{\ast}=J$, $d^{2}_{P}y^{\ast}=\nabla J=K$, so $D_{2}=\Psi_{2}(J,K)$. If $D_{2}$ is
affine-null then $y^{\ast}(P)=AP+b$ gives $J=A$, $K=0$, $D_{2}=0$; hence $D_{2}$ vanishes on all of
$\{K=0\}$, so any part surviving there requires $K$: the nontrivial second-order content is carried
by $K$. We do \emph{not} claim $D_{2}$ factors through $K$ alone: the relative curvature
$\mathrm{relK}=\|K[v]\|/\|Jv\|$ depends on both $J$ and $K$, and is affine-null precisely because its
numerator vanishes with $K$.
\emph{(iii) Path level.} Path-locality makes $D(\gamma)$ depend on $y^{\ast}$ along $\gamma$ only
through $\dot y=J\dot\gamma$, the integral-curve equation of $\omega=J\,dP$, with solution
$y(1)=y(0)+\int_{\gamma}\omega$. Reparametrization invariance discards the traversal speed, leaving
dependence only on the oriented transport
$\mathcal T^{\omega}_{\gamma}:y(0)\mapsto y(0)+\int_{\gamma}\omega$; for a closed loop the identity
transport is $y^{\ast}(P_{0})$, so any loop-closure diagnostic factors through
$\mathrm{Hol}_{\omega}(\gamma)=\mathcal T^{\omega}_{\gamma}-I$. The loop-closure restriction is
essential: a general path-local, reparametrization-invariant functional such as the embedded arc
length $\int_{\gamma}\|\dot y\|$ depends on the whole trajectory, not on
$\mathcal T^{\omega}_{\gamma}$ alone.\hfill$\square$

\smallskip\noindent\emph{Scope (the two limits the result turns on).} First, an affine-null
second-order diagnostic need not factor through $K$ \emph{alone}: the framework's own relative
curvature $\mathrm{relK}=\|K[v]\|/\|Jv\|$ depends on both $J$ and $K$, and is affine-null precisely
because its numerator vanishes with $K$. What (ii) establishes is that the \emph{nontrivial}
second-order content is carried by $K$, not that $J$ drops out. Second, the loop-closure restriction
in (iii) is essential, as the arc-length counterexample above shows. Finally, equivariance is taken
under Euclidean isometries of the embedding, matching the fixed $\ell_{2}$ metric the readings use;
covariance under a general diffeomorphism would require replacing $K=D^{2}_{P}y^{\ast}$ with the
covariant Hessian $\nabla\,dy^{\ast}$, and is not claimed here.

\subsection{Corollary~1 (structural independence)}
By the reading ladder above, a linear objective gives $K\equiv0,\ \mathrm{Hol}\equiv0$; a strongly
convex non-quadratic objective can give $K\neq0$ with $\mathrm{Hol}\equiv0$; and a non-convex
objective can give both non-zero. Already at this level the map curvature does not determine the
holonomy: the second row has $K\neq0$ and $\mathrm{Hol}\equiv0$, and the third can have $K$ of any
size with $\mathrm{Hol}\neq0$, so a single-valued $\Phi$ with $\mathrm{Hol}=\Phi(K)$ would have to
send one curvature value to two holonomies.

The witnesses in Table~\ref{tab:ladder} instantiate this on identical data. Compare the autoencoder
(an explicit map: $\omega=d\Phi$ is exact, so $\mathrm{Hol}\equiv0$ \emph{structurally}, for
\emph{every} loop) with t-SNE (non-convex argmin, $\mathrm{Hol}=0.15$). Their curvatures are of the
same order and their across-seed ranges overlap ($\mathrm{relK}\in[1.5,4.6]$ for the autoencoder,
$[2.7,4.8]$ for t-SNE over seeds $0$--$2$), so there are curvature values realized by both
constructions at which the holonomy is exactly $0$ and $0.15$ respectively. Hence no single-valued
$\mathrm{Hol}=\Phi(\mathrm{relK})$ exists, and the two readings are functionally
independent.\hfill$\square$

\subsection{Theorem~1 (iv): minimality of the integral reading}\label{app:minimality}
Call $L_S(\omega)=\Phi\!\big(j^{m}_{P_1}\omega,\dots,j^{m}_{P_N}\omega\big)$ a \emph{finite-order
local diagnostic} on a finite sample $S=\{P_1,\dots,P_N\}\subset\mathcal U_0$: any quantity built from
the $m$-jets of the connection at finitely many points. Fix $S$, an order $m$, and a loop $\ell$
avoiding $S$. Since $S$ is finite and $\ell\setminus S$ is open, $\ell$ has an arc $A$ bounded away
from every $P_i$; pick a bump $b\in C^\infty(\mathcal U_0)$ supported in a tubular neighbourhood of
$A$ that vanishes to infinite order on a neighbourhood of each $P_i$ (such $b$ exists because $A$ is
disjoint from the finite set $S$). Take $f_0$ strongly convex in $y$ along $\ell$, with unique
nondegenerate branch $y_0^{\ast}(P)$; by P2\,(i) its transport closes, $\mathrm{Hol}_{\omega_0}(\ell)=0$.
Set
\[
f_1(P,y)=f_0(P,y)+\eta\,b(P)\,q(y),
\]
with $q\in C^\infty$ and $\eta$ chosen so that on the arc $A$ the added term drives the tracked
minimum into a fold (saddle-node): $\lambda_{\min}(H_{yy})\!\to\!0$ and $y_0^{\ast}$ collides with an
emerging critical point, so the branch cannot be continued single-valuedly across $A$ (the fold
normal form $\tfrac13y^3-\lambda(P)y$ localized by $b$ realizes this). Because $b$ and all its
derivatives vanish near each $P_i$, $f_1\equiv f_0$ to infinite order there, so the two induced
branches share every jet at each $P_i$: $j^{m}_{P_i}\omega_1=j^{m}_{P_i}\omega_0$, hence
$L_S(\omega_1)=L_S(\omega_0)$ for \emph{every} finite-order local diagnostic. Yet the branch of $f_0$
continues around $\ell$ and returns (branch-distance $0$) while the branch of $f_1$ meets the fold on
$A$ and fails to continue (nonzero branch-distance / continuation failure, P2\,(ii)): two genuine
smooth objectives with identical finite-order local data on $S$ have different branch-continuation
around $\ell$. No finite-order local diagnostic sampled on $S$ is therefore a complete detector of
path dependence. Since two paths sharing endpoints are path-equivalent iff the loop they form
continues trivially, this branch-continuation obstruction is the minimal one to path
independence.\hfill$\square$

\emph{Scope.} The witnesses are two genuine smooth objectives, so the non-identifiability holds
within the DR-induced class the framework reads, not merely at the level of abstract one-forms:
finite discrete local sampling cannot determine branch events (saddle-node, branch loss, basin
reconnection) that occur in an unsampled path region. It does not assert that a diagnostic reading
local data \emph{continuously} along the whole path is blind --- such a diagnostic has, by
definition, already become an integral reading. The non-zero t-SNE branch-distance of
Sec.~\ref{sec:holonomy} (P2) instantiates the separation on real data, along loops where the sampled
local jets are unremarkable.

\subsection{Theorem~2 (spectral divergence law) and Proposition~P3 (complementarity)}
The \emph{Divergence} and \emph{Rate ladder} paragraphs below prove Theorem~2; the
\emph{Complementarity} paragraph proves P3.
\emph{Divergence.} On $D_{\mathrm{loc}}$, $\lambda_{\min}\to0$, so $\lambda_{\min}^{-1}\to\infty$ by
definition; this is the divergence of Liu et al.'s score, predicted by the rate law rather than
posited. The score itself is the well-posedness gate of Sec.~\ref{sec:level0}, which does not factor
through any reading of $\Phi$.
\emph{Rate ladder (upper bounds).} Since $H_{yy}\succ0$ off $D_{\mathrm{loc}}$,
$\|H_{yy}^{-1}\|_{2}=\lambda_{\min}^{-1}$, so
$\|J\|_{2}\le\|H_{yy}^{-1}\|_{2}\|H_{yx}\|_{2}=\|H_{yx}\|_{2}\,\lambda_{\min}^{-1}$, with equality
when $H_{yx}$ maps onto the $\lambda_{\min}$-eigenvector. The curvature identity of
Sec.~\ref{sec:curvature} is $K[v]=-H_{yy}^{-1}a$ with $a$ the second directional derivative of
$\nabla_{y}f$ along $(v,Jv)$; bounding $\|a\|\le M_{3}(1+\|J\|)^{2}$ with $M_{3}=\sup\|\nabla^{3}f\|$
gives $\|K[v]\|\le\lambda_{\min}^{-1}M_{3}(1+M_{2}\lambda_{\min}^{-1})^{2}
=M_{3}M_{2}^{2}\lambda_{\min}^{-3}+O(\lambda_{\min}^{-2})$, i.e.\ $O(\lambda_{\min}^{-3})$, tight
under the same alignment. These are worst-case bounds; on a real path they are attained only when
the soft mode is excited (i.e.\ $\|u^{\top}H_{yx}\|$ is non-negligible for $u$ the $\lambda_{\min}$
eigenvector), matching the measured slopes $-0.99/-3.00$ in the excited synthetic family.
\emph{Complementarity.} By construction $D_{\mathrm{loc}}\subseteq D$. \emph{(i)} At a point of
$\{\text{argmin non-unique}\}\setminus D_{\mathrm{loc}}$ each competing minimizer is nondegenerate,
so $\lambda_{\min}$ is bounded away from $0$ and the $\lambda_{\min}$-driven conditioning magnitudes
($\|J\|$, $\lambda_{\min}^{-1}$, $\|K\|$) are finite: they register nothing there, whereas the holonomy
of P2 --- non-zero when the loop's induced transport meets the degeneracy locus $\Sigma$ or its
terminal snap crosses a basin boundary --- and the integrated flow's endpoint do register the global
basin-swap of $\{\text{argmin non-unique}\}$. In this conditioning sense the differential reach lies
inside the integral reach, \emph{local}\,$\subset$\,\emph{global}. \emph{(ii)} The reverse inclusion
fails: the interior integrability form $F$ of Cor.~3 is a purely local, second-order reading that is
non-zero on well-posed metric MDS --- exactly where the loop holonomy is zero --- so the endpoint
transport does not recover it. Neither reach therefore contains the other; the conditioning
magnitudes and the loop obstruction are complementary detectors.\hfill$\square$

\begin{figure}[t]\centering
\includegraphics[width=\columnwidth]{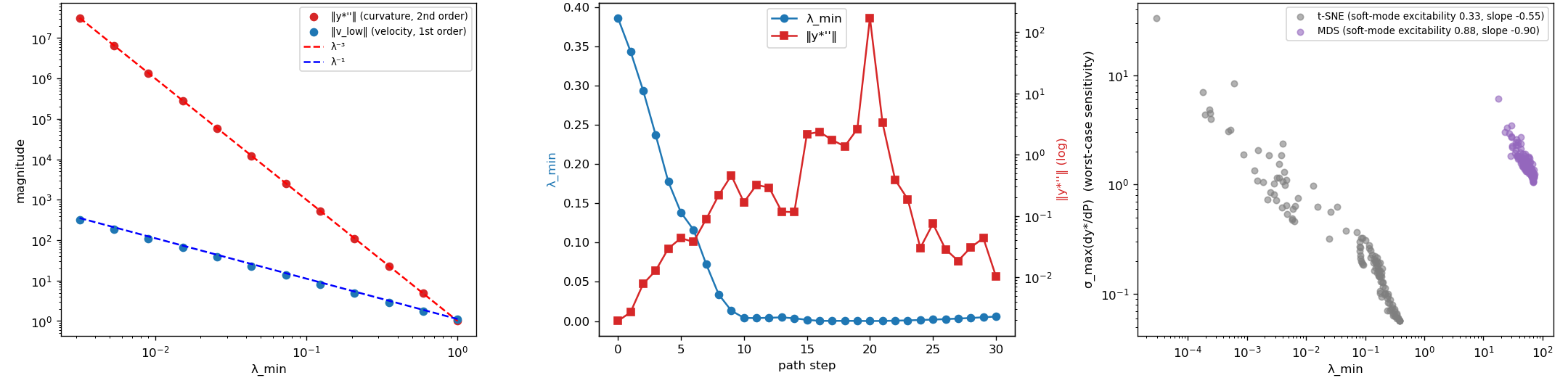}
\caption{Empirical verification of the spectral divergence law (Theorem~2), and why its exponents are worst-case upper bounds.
\emph{(1)} On a controlled synthetic family in which only $\lambda_{\min}$ varies and the soft mode
is fully excited, the first- and second-order operator norms diverge at measured slopes $-0.99$ and
$-3.00$ ($R^2=1.00$), matching the predicted $\lambda_{\min}^{-1}$ and $\lambda_{\min}^{-3}$: the
worst-case exponents are achievable. \emph{(2)} On a real digits t-SNE path the curvature peak still
co-locates with the $\lambda_{\min}$ dip, but the sensitivity grows only as $\lambda_{\min}^{-0.55}$:
the soft mode is weakly excited, so the $\lambda_{\min}^{-3}$ bound is loose. \emph{(3)} The
separating quantity is the soft-mode excitability $\|u^\top H_{yx}\|/\|H_{yx}\|$ ($u$ the
$\lambda_{\min}$ eigenvector): metric MDS excites it (bound tight) while t-SNE decouples (bound
loose).}
\label{fig:rateladder}
\end{figure}

\subsection{Proposition~P1 (glyph-fidelity $s^{2}$ certificate)}
Fix $P$ and a unit direction $v$, and let $\varphi(s)=y^{\ast}(P+sv)$ on an interval on which
$H_{yy}$ is nonsingular; $\varphi\in C^{2}$ by the IFT and $f\in C^{3}$. Differentiating the
optimality identity $\nabla_{y}f(P+sv,\varphi(s))\equiv0$ once gives, by the chain rule,
$\varphi'(s)=J(P+sv)\,v$, hence $\varphi'(0)=Jv$; differentiating again along the tangent
$(v,\varphi'(s))$ gives $\varphi''(s)=K[v](P+sv)$. Taylor's theorem with integral remainder
yields $\varphi(s)=\varphi(0)+s\varphi'(0)+\int_{0}^{s}(s-t)\varphi''(t)\,dt$, so
\begin{equation*}
y^{\ast}(P+sv)-(y^{\ast}+sJv)=\int_{0}^{s}(s-t)\,K[v](P+tv)\,dt.
\end{equation*}
Taking norms and bounding $\int_{0}^{s}(s-t)\,dt=\tfrac12 s^{2}$,
$\|y^{\ast}(P+sv)-(y^{\ast}+sJv)\|\le\tfrac12 s^{2}\sup_{t\in[0,s]}\|K[v](P+tv)\|$; expanding
$K[v](P+tv)=K[v](P)+O(t)$ gives the stated $\tfrac12 s^{2}\|K[v,v]\|+O(s^{3})$.\hfill$\square$

\begin{figure}[t]\centering
\includegraphics[width=\columnwidth]{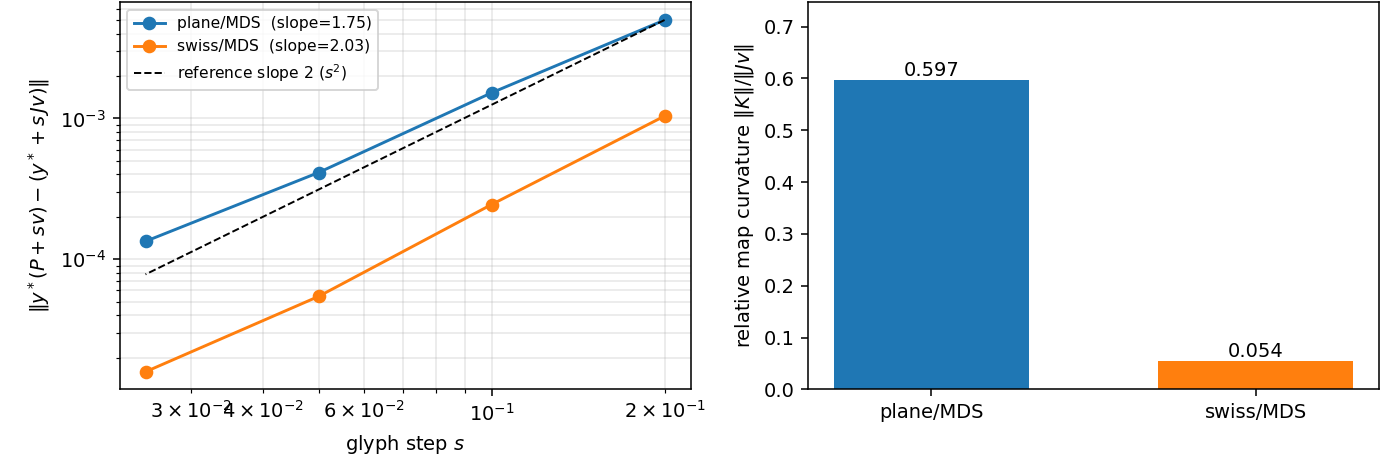}
\caption{Empirical verification of the P1 glyph-fidelity certificate, on two MDS embeddings.
\emph{Left:} the measured finite-step
glyph error scales as $s^2$ (both cases track the reference
slope-2 line), confirming P1: the glyph's linear extrapolation degrades
quadratically in the step. \emph{Right:} the relative map curvature is larger on
the flat plane than on the curved Swiss roll --- on these two
testbeds, curvature tracks the objective rather than the data manifold.}
\label{fig:curv}
\end{figure}

\subsection{Proposition~P2 (integrability dichotomy)}
Let $g(P,y)=\nabla_{y}f(P,y)$, and on $P\setminus\Sigma$ (with $\Sigma$ the caustic
$\{\det H_{yy}=0\}$ projected to $P$-space) let $y^{\ast}(P)$ be a nondegenerate minimum branch,
$C^{2}$ by the IFT. Differentiating $g(P,y^{\ast}(P))\equiv0$ gives $dy^{\ast}=J\,dP=\omega$, so
on the domain of a single smooth branch the induced-transport one-form $\omega$ \emph{is exact}
($\omega=dy^{\ast}$), hence closed. Consequently, for any loop $\gamma$ that bounds a disk
$D\subset P\setminus\Sigma$ on which the branch $y^{\ast}$ persists,
$\oint_{\gamma}\omega=\oint_{\gamma}dy^{\ast}=0$: the continuous flow returns exactly and the
branch-distance is $0$.

\emph{(i) Exact-transport direction (certified).} If $f(P,\cdot)$ is strongly convex in $y$ on a
neighborhood of a filling disk of $\gamma$ ($H_{yy}\succ0$, unique minimum), the single min-branch
is defined on all of $D$, so by the above the branch-distance is identically $0$, with no snap
invoked. For metric MDS, $H_{yy}\succ0$ is (near-)constant along the tested loops, so $\omega$ is
exact and holonomy $\equiv0$; this is the MDS baseline and the only direction we certify.

\emph{(ii) Non-exact direction (diagnostic, not topological).} A non-zero converged
branch-distance certifies that no such filling disk exists: the loop's tracked branch cannot be
continued single-valuedly over any disk it bounds. Two non-topological mechanisms produce this,
distinguished by the spectral certificate $[\,\|v_{\mathrm{low}}\|,\lambda_{\min},\Delta f\,]$:
the flow passing near/through $\Sigma$, where $\|H_{yy}^{-1}\|\to\infty$ and the tracked minimum
degenerates; and/or the terminal snap landing in a competing basin (a global-argmin
reconfiguration at the Maxwell set, to which the pure-RK4 local-branch flow is otherwise
insensitive). In both cases the branch-distance is a continuous, loop- and basepoint-dependent
magnitude, not a quantized invariant.\hfill$\square$

\emph{Remark (scope: a graded magnitude, not an invariant).} We claim no topological (covering-monodromy)
reading. Since $\Sigma$ is generically codimension one ($\det H_{yy}=0$, a single scalar equation),
a $\Sigma$-avoiding loop bounds a disk in $P\setminus\Sigma$ and transports exactly by
direction~(i); any non-zero value therefore comes from the analytic mechanisms of direction~(ii),
so the branch-distance is a continuous conditioning magnitude read as order and separation, which is
the appropriate (and strictly more informative) form for a diagnostic that must grade the
degree of ill-posedness.

\subsection{Corollary~3 (differential--integral decoupling)}
One might hope the loop holonomy of Sec.~\ref{sec:holonomy} is the integral of a local curvature
2-form, hence a Stokes/Ambrose--Singer~\cite{ref:ambrosesinger} invariant. It is not, and the obstruction is constructive.
The induced submersion $y^{\ast}:\mathcal U_{0}\to\mathcal Y$ does carry a genuinely non-flat
connection (the Ehresmann connection of its horizontal distribution
$\mathcal H=(\ker J)^{\perp}=\mathrm{span}\{g_{a}\}$, $g_{a}=\nabla_{P}y^{\ast}_{a}$), whose
O'Neill~\cite{ref:oneill} integrability $2$-form is
\begin{equation*}
F_{ab}=\mathcal V\,[g_{a},g_{b}]=\mathcal V\big(K_{b}g_{a}-K_{a}g_{b}\big),\qquad \mathcal V=I-J^{+}J,
\end{equation*}
where $K_{a}=\mathrm{Hess}(y^{\ast}_{a})$ is the $a$-th slice of $K$ and the bracket identity
$[g_{a},g_{b}]=K_{b}g_{a}-K_{a}g_{b}$ follows from the symmetry of each $K_{a}$. Thus $F$ is a
function of $(J,K)$ and is generically non-zero on a curved induced map, while $F$ is machine-zero for a linear map
($K=0$) and, degenerately, for a flat data subspace on which $\mathcal H$ is a constant plane. But
$F$ is a purely \emph{interior}, second-order object, and it is non-zero exactly on well-posed
metric MDS, where the \emph{loop} holonomy of Sec.~\ref{sec:holonomy} is zero. The two have
disjoint non-zero patterns, so the loop holonomy is not the integral of $F$, nor of any curvature
built from the local jet; it is the branch obstruction of the flat point-transport
$\omega=dy^{\ast}$ across the caustic (P2), Aharonov--Bohm-\emph{like}~\cite{ref:aharonovbohm} by analogy but, since
$\Sigma$ is codimension one, neither a local-curvature flux nor a topological invariant. This is the constructive
form of the map-curvature/connection-curvature distinction of Sec.~\ref{sec:curvature}.

\subsection{Proposition~P5 (finite-scale neighbourhood preservation) and relational irreducibility}\label{app:finitescale}
\emph{Lemma (finite-scale distortion).} Let $v=x_{j}-x_{i}$ with $\|v\|=d_{ij}\le R$. As in the P1
proof, Taylor with integral remainder gives
$\Phi(x_{i}+v)-\Phi(x_{i})=J_{i}v+\mathcal R_{i}(v)$ with
$\|\mathcal R_{i}(v)\|\le\tfrac12\sup_{B_{R}}\|\nabla_{P}J\|\,\|v\|^{2}=\tfrac12 M_{i}(R)\|v\|^{2}$,
$M_{i}(R)=\sup_{B_{R}}\|K\|$. On the data-manifold tangent, $Q_{i}$ is an isometry
($\|Q_{i}v\|=\|v\|$), so by the reverse triangle inequality and the operator-norm bound,
$\bigl|\,\|J_{i}v\|-\alpha_{i}\|v\|\,\bigr|=\bigl|\,\|J_{i}v\|-\|\alpha_{i}Q_{i}v\|\,\bigr|\le
\|(J_{i}-\alpha_{i}Q_{i})v\|\le\delta_{i}\|v\|$. Since $\hat d_{ij}=\|J_{i}v+\mathcal R_{i}(v)\|$,
\[
\begin{aligned}
\bigl|\hat d_{ij}-\alpha_{i}d_{ij}\bigr|
&\le\bigl|\,\|J_{i}v\|-\alpha_{i}\|v\|\,\bigr|+\|\mathcal R_{i}(v)\|\\
&\le\delta_{i}\|v\|+\tfrac12 M_{i}(R)\|v\|^{2}\\
&\le\delta_{i}R+\tfrac12 M_{i}(R)R^{2}=\epsilon_{i}(R).
\end{aligned}
\]
(For densely sampled or flat data the chord $v$ coincides with the tangent step; on a curved data
manifold one replaces $v$ by the geodesic and adds the data manifold's own second-fundamental-form
term, cf.\ the scope note of Sec.~\ref{sec:inverse}.)\hfill$\square$

\smallskip\noindent\emph{Proposition~P5.} For $j\in N_{k}^{X}(i)$ we have $d_{ij}\le d_{i,(k)}$, so the
Lemma gives $\hat d_{ij}\le\alpha_{i}d_{i,(k)}+\epsilon_{i}(R)$; for $l\notin N_{k}^{X}(i)$ we have
$d_{il}\ge d_{i,(k+1)}$, so $\hat d_{il}\ge\alpha_{i}d_{i,(k+1)}-\epsilon_{i}(R)$. When
$\alpha_{i}\Delta_{i,k}=\alpha_{i}(d_{i,(k+1)}-d_{i,(k)})>2\epsilon_{i}(R)$, the two bounds separate,
$\alpha_{i}d_{i,(k)}+\epsilon_{i}(R)<\alpha_{i}d_{i,(k+1)}-\epsilon_{i}(R)$, hence
$\hat d_{ij}<\hat d_{il}$ for every in/out pair: the low- and high-d $k$-neighbourhoods
coincide, $N_{k}^{X}(i)=N_{k}^{Y}(i)$, so $T_{k}(i)=C_{k}(i)=1$. For the localization, a member of the
symmetric difference is some $j$ whose induced distance crosses to the wrong side of $\hat d_{i,(k)}$;
by the Lemma a crossing forces $|d_{ij}-d_{i,(k)}|\le 2\epsilon_{i}(R)/\alpha_{i}$, so
$N_{k}^{X}(i)\,\triangle\,N_{k}^{Y}(i)\subseteq\{j:|d_{ij}-d_{i,(k)}|\le 2\epsilon_{i}(R)/\alpha_{i}\}$,
and summing the rank penalties over this band bounds the trustworthiness/continuity loss. The
hypothesis that all candidates lie within $R$ is the no-nonlocal-intrusion condition; where a distant
point is contracted into the neighbourhood it must be added to the band by hand.\hfill$\square$

\smallskip\noindent\emph{Remark (local jets do not determine the relational reading).} Fix a query
$x_{0}$, an order $m\ge0$, an open neighbourhood $U\ni x_{0}$, and $k\ge1$. There are a finite sample
$X$ and two smooth embeddings $\Phi_{0},\Phi_{1}$ with $\Phi_{0}|_{U}=\Phi_{1}|_{U}$ --- hence equal
$m$-jets at $x_{0}$, indeed identical differential readings of every order there --- yet
$T_{k}^{\Phi_{0}}(x_{0})\neq T_{k}^{\Phi_{1}}(x_{0})$. Take a true high-d near neighbour
$x_{a}$ and a far point $x_{b}$, and $\Phi_{0}$ with
$\|\Phi_{0}(x_{a})-\Phi_{0}(x_{0})\|<\|\Phi_{0}(x_{b})-\Phi_{0}(x_{0})\|$. Choose a smooth bump $\psi$
supported outside $U$ and covering $x_{b}$, and set $\Phi_{1}=\Phi_{0}+a\psi$ with $a$ chosen so that
$\|\Phi_{1}(x_{b})-\Phi_{1}(x_{0})\|<\|\Phi_{1}(x_{a})-\Phi_{1}(x_{0})\|$. Since $\psi|_{U}=0$ we have
$\Phi_{1}|_{U}=\Phi_{0}|_{U}$, so every local jet at $x_{0}$ is unchanged, while $x_{b}$ has become a
false neighbour and $T_{k}$ has moved. No function of the finite-order local jet can therefore
reproduce $T_{k}$: the relational reading's irreducible input is the sample configuration --- the
positions of the other points, the scale $k$, and the margin $\Delta$ that P5 isolates. This mirrors
the minimality of the integral reading (Thm.~1(iv)): the relational and integral readings each exceed
pointwise differential data, in complementary ways (finite-sample rank versus path transport).\hfill$\square$

\subsection{The reading ladder}
If the embedding is given by an explicit differentiable map $\Phi$ (no objective; $\Sigma=\emptyset$),
the connection $J=\partial\Phi/\partial P$ is a global Jacobian, so the holonomy is $\equiv0$
identically while the map curvature $K$ can be non-zero: the differential reading is alive and the
integral reading is vacuous. If $f(\cdot,y)$ is quadratic in $y$ with a $P$-independent Hessian (linear DR, e.g.\
PCA's $\|x-Wy\|^{2}$), then $J=W^{\top}$ is constant, so $K=\nabla_{P}J\equiv0$, and the unique
minimizer with $H_{yy}\succ0$ everywhere gives holonomy $\equiv0$: both new readings are
machine-zero. If $f(P,\cdot)$ is strongly convex but not quadratic, the differential curvature $K$
can be non-zero while the minimum stays unique, so $D_{\mathrm{loc}}=\emptyset$ and the integral
holonomy $\equiv0$. If $f(P,\cdot)$ is non-convex, both $D_{\mathrm{loc}}$ and
$\{\text{argmin non-unique}\}$ can be non-empty and both new readings can be non-trivial. Thus the
set of readings that can be non-zero grows monotonically with the objective's structural
complexity.\hfill$\square$

\emph{Remark (the exactly-integrable corner: a closed-form trajectory that need not be flat).}
The ladder classifies which readings vanish; here is the corner in which the induced integral
itself is elementary. Let $f$ be quadratic in $y$ with a positive-definite Hessian,
$f(P,y)=\tfrac12 y^{\top}A(P)\,y+b(P)^{\top}y+c(P)$ with $A(P)\succ0$. The optimality condition
$A(P)\,y+b(P)=0$ is linear in $y$, so the induced map is the closed form
\begin{equation*}
y^{\ast}(P)=-A(P)^{-1}b(P),
\end{equation*}
and the corrected trajectory along any high-d path $P(t)$ is
$y(t)=-A(P(t))^{-1}b(P(t))$ --- a rational map evaluated pointwise, with no integration performed.
Since $H_{yy}=A(P)\succ0$ everywhere the caustic is empty ($\Sigma=\varnothing$), the global
min-branch is single-valued, and $\omega=dy^{\ast}$ is exact, so $\mathrm{Hol}\equiv0$ (P2\,(i)).
The connection is $J=-A^{-1}\big((\partial_{P}A)\,y^{\ast}+\partial_{P}b\big)$.

The map is flat exactly when $y^{\ast}(P)=-A(P)^{-1}b(P)$ is affine in $P$: the canonical flat
instance has a constant Hessian $A$ and $b$ affine, so $J$ is constant and $K\equiv0$ --- the
linear-DR row already in the ladder (PCA). Otherwise, whenever $A$ varies with $P$ or $b$ is
nonlinear, $y^{\ast}$ is genuinely nonlinear in $P$, so $J$ varies and $K=\nabla_{P}^{2}y^{\ast}$ is
generically non-zero, yet the trajectory is still closed form. This is the one cell the
``quadratic $\Rightarrow$ flat'' reading omits: \emph{a closed-form integral does not imply a flat
one}, and the curvature here comes from the nonlinearity of $y^{\ast}(P)$, not from any degeneracy
($\Sigma$ stays empty). A concrete member is a kernel-weighted
(Nadaraya--Watson-type) OOS placement against fixed anchors $\{\eta_{i}\}$:
$f(P,y)=\sum_{i}w_{i}(P)\,\|y-\eta_{i}\|^{2}$ gives the barycentric closed form
$y^{\ast}(P)=\big(\sum_{i}w_{i}(P)\eta_{i}\big)\big/\big(\sum_{i}w_{i}(P)\big)$, which bends with the
affinities $w_{i}(P)$ ($K\neq0$) while closing exactly ($\mathrm{Hol}\equiv0$). The corner thus sits
strictly between linear DR (closed form, flat) and the strongly-convex non-quadratic case (no closed
form, curved): closed-form transport with live differential curvature.\hfill$\square$

\subsection{Protocol robustness of the integral reading}\label{app:protocol}
The integral reading's verdict --- t-SNE non-integrable (hardened branch-distance $\ge0.10$) and
metric MDS integrable ($\le0.02$) --- is invariant to the numerical protocol on the controlled
3-cluster loop where the phenomenon is drawn. Varying each knob in turn while holding the rest at
baseline (Table~\ref{tab:protocol}), the verdict holds on all $24$ protocol variants across six axes
and on all five embedding seeds; on the seed axis the hardened t-SNE branch-distance is $0.154$ to
five significant figures. What is \emph{not} invariant is the raw loop-closure gap: its t-SNE/MDS
separation ratio ranges from $4.4$ to $3.7\times10^{8}$ and its raw magnitude by ${\sim}23\times$ over
the same sweep. This contrast is exactly why the reading is reported as the hardened branch-distance
read for order and separation (Sec.~\ref{sec:holonomy}), not as the raw gap. Two caveats stay within
the scope already stated: the branch-distance is basepoint-dependent (the Remark above), so at some
basepoints the t-SNE loop reads integrable (branch ${\sim}10^{-5}$) --- consistent with a
non-integrability certificate that is existential over loops, while MDS stays integrable at every
basepoint; and the branch-distance can overshoot at coarse step size (a basin jump), which is why the
fine substeps tail is the gated quantity.

\begin{figure}[t]\centering
\includegraphics[width=\columnwidth]{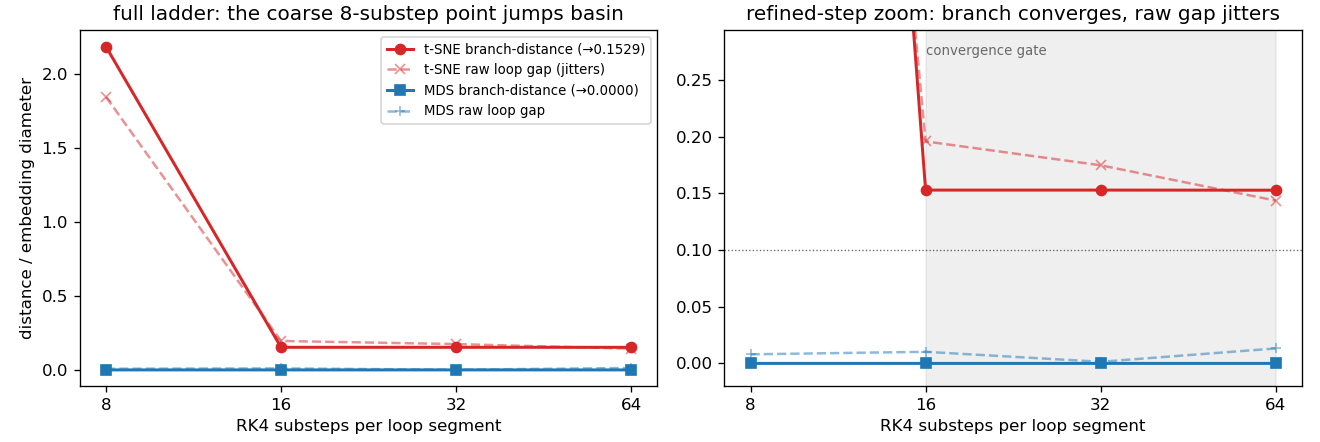}
\caption{Hardening the holonomy into a step-refinement--stable reading. As the RK4 substeps per
loop segment increase, the t-SNE \emph{branch-distance} (solid) converges to a stable value
($\approx0.15$), whereas the \emph{raw} loop-closure gap (dashed) keeps jittering, which is
why we read the branch-distance, not the raw gap. \emph{Left:} the full ladder; the coarse
8-substep point lands in a wrong basin, a discretization artefact the convergence gate excludes.
\emph{Right:} the refined-step zoom over the gate, where the t-SNE branch-distance is flat and
metric MDS stays at zero. The separation is a step-refinement--stable diagnostic of exact
(well-posed, MDS) versus non-exact (ill-posed, t-SNE) induced transport.}
\label{fig:holo}
\end{figure}

\begin{table}[t]\centering\small
\caption{Protocol robustness of the integral-reading verdict on the controlled 3-cluster loop
(App.~\ref{app:protocol}). Each axis is swept with the others held at baseline; ``verdict'' counts
variants on which t-SNE reads non-integrable \emph{and} MDS integrable. The raw loop-closure gap
varies by orders over the same sweep (text); the hardened branch-distance does not.}
\label{tab:protocol}
\begin{tabular}{@{}lll@{}}
\toprule
Axis & Swept range & Verdict\\
\midrule
Step size (substeps) & $2$--$32$ & $5/5$\\
Tikhonov $\lambda_{\mathrm{reg}}$ & $10^{-4}$--$10^{-1}$ & $4/4$\\
Terminal snap rule & restarts $1$--$4$, steps $200$--$800$ & $6/6$\\
Loop radius $R$ & $2.5$--$4.0$ & $3/3$\\
Loop vertices $M$ & $32$--$64$ & $3/3$\\
Perplexity & $15$--$50$ & $3/3$\\
Embedding seed & $\{0,1,2,3,42\}$ & $5/5$\\
\bottomrule
\end{tabular}
\end{table}

\section{Instrument sanity checks: OOS placement and cost}\label{app:sanity}
The two readings of Secs.~\ref{sec:differential}--\ref{sec:integral} are
properties of the induced connection and hold whatever the quality of any particular implementation
of them. What the two checks below establish is that our implementation is a usable instrument --- that a machinery
which reads placements can itself place a point competitively, and that integrating the flow is
affordable at the rate an analyst works.

\emph{OOS neighborhood preservation.} A framework that reads placements should itself
place points competitively, so we compare OOS projection quality (kNN recall and
trustworthiness) against re-embedding-with-Procrustes, UMAP-transform, and nearest-anchor
baselines across MDS, t-SNE, Isomap, and UMAP. On well-posed embeddings our OOS placement matches
re-embedding and far exceeds the naive baseline, and on a COIL-20 t-SNE map it beats re-embedding
outright (0.796 vs 0.513). It underperforms in a flat, uniform plane, the regime that t-SNE
fractures into fabricated clusters, where our integration falls below the nearest-anchor baseline
(0.521 vs 0.729). It is the ill-posed regime the framework's readings are built to expose, where the
induced geometry is genuinely degenerate.

\emph{Cost.} Reading the integral geometry over a whole map integrates the induced flow along many
high-d routes, over many frames and many seeds; the framework is tractable because that
integration \emph{amortizes}. On the FashionMNIST morph of
Fig.~\ref{fig:tear} it runs $\approx2.6\times$ cheaper in objective evaluations than point-wise full
re-optimization along the same path while remaining faithful, and $\approx3.6\times$ on the
perturbation paths of Sec.~\ref{sec:adv} (App.~\ref{app:advpath}). Both are
constant-factor savings; we report no wall-clock and no scaling in $n$, and
the morph comparison inherits an under-correction floor that bounds how far the saving can be pushed.
This makes the
perturbation-path scenario practical at the frame rate of a morph, which is the claim we need, not
scalability to a whole map.

\section{Perturbation-path amortization across source geometries}\label{app:advpath}
\begin{figure*}[t]\centering
\includegraphics[width=\textwidth]{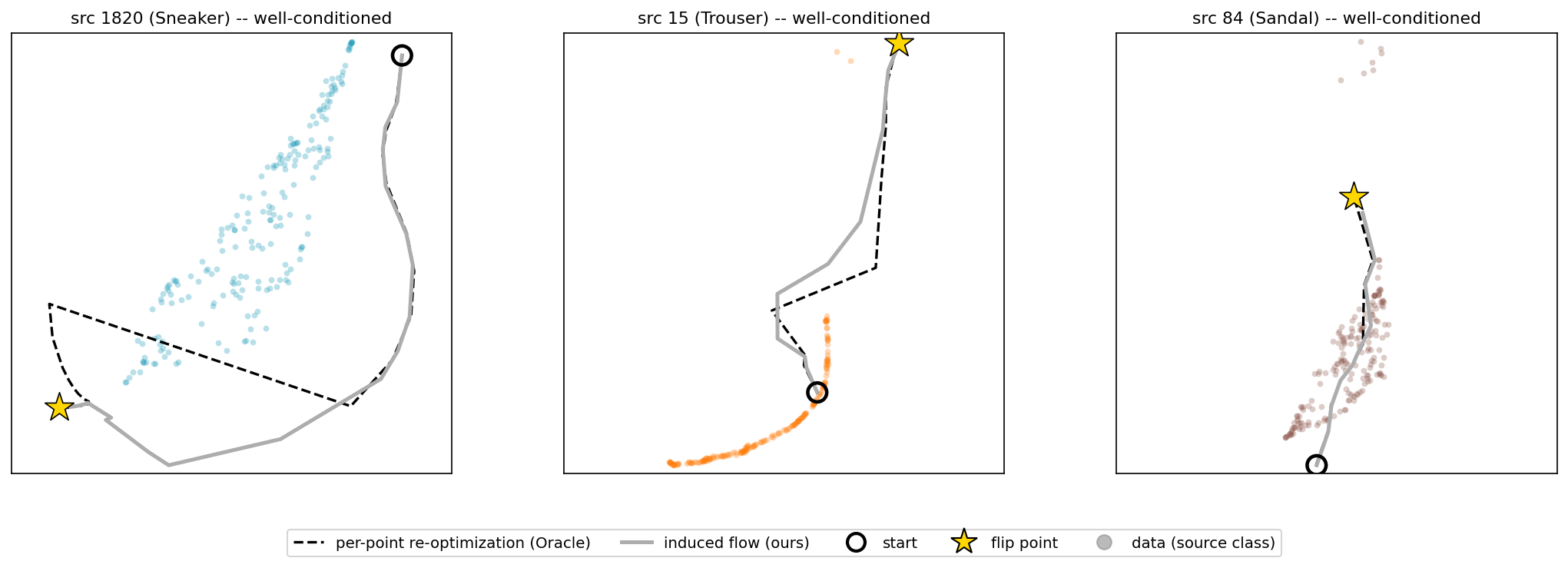}
\caption{Perturbation-path amortization on all three shipped sources (the main-text
Fig.~\ref{fig:advpath} shows src~84 alone). Each panel plots the induced flow (grey) against the
per-point re-optimization Oracle (dashed), from start ($\circ$) to classifier-flip placement ($\star$),
over the source-class cluster. Panels show the eight-seed pool's median-draw values; the matching and
amortization figures are in the text above.}
\label{fig:advpath3}
\end{figure*}

The single-source panel of Fig.~\ref{fig:advpath} is a legible instance of a result we report over
three source classes chosen for distinct path geometries: skirting a cluster edge (src~1820, Sneaker),
extending out of a cluster (src~15, Trouser), and passing through one (src~84, Sandal).
Fig.~\ref{fig:advpath3} shows all three. On each, the flow matches the Oracle to a median
$0.008\cdot$diam (per-source $\le0.012$) at $\Delta f=5\times10^{-4}$ while amortizing the optimum at
$0.24$--$0.42\times$ its objective evaluations ($\approx3.6\times$ fewer). The amortization, and the
flow's smoothness where per-frame re-optimization corners between adjacent minima, thus hold across
the three geometries, not only on the src-84 path shown in the main text; all three are
well-conditioned by design ($\lambda_{\min}$ never collapses).

\section{The rotation-cycle exhibit (COIL-20)}\label{app:coil}
This exhibit expands the pointer in Sec.~\ref{sec:holonomy}. COIL-20~\cite{ref:coil20} is the canonical dataset in
which each object's rotation frames trace a closed loop, and an analyst reading a t-SNE map naturally
concludes that each ring is a faithful, closed viewpoint cycle. Holding an object out and integrating
its rotation loop through the induced flow asks a question the static glyph and the point-level score
structurally cannot: does the map return the closed viewpoint cycle to itself? Because every
high-d rotation loop is closed by construction, any failure to close is a pure map
artifact --- a path-inconsistency invisible to the eye and to both predecessors~\cite{ref:bian,ref:liu}.
We present this as a \emph{qualitative} concept exhibit. On this real, strongly multimodal t-SNE
landscape the per-object loop gap is not stable under integration-step refinement (the
branch-distance hardening of Sec.~\ref{sec:holonomy} does not converge across resolutions here, as it
does on the controlled synthetic), so we make no per-object ranking or magnitude claim and read the
vignette only as an illustration of the class of path-inconsistency the integral reading can pose and
the others cannot.

\begin{figure*}[t]\centering
\includegraphics[width=\textwidth]{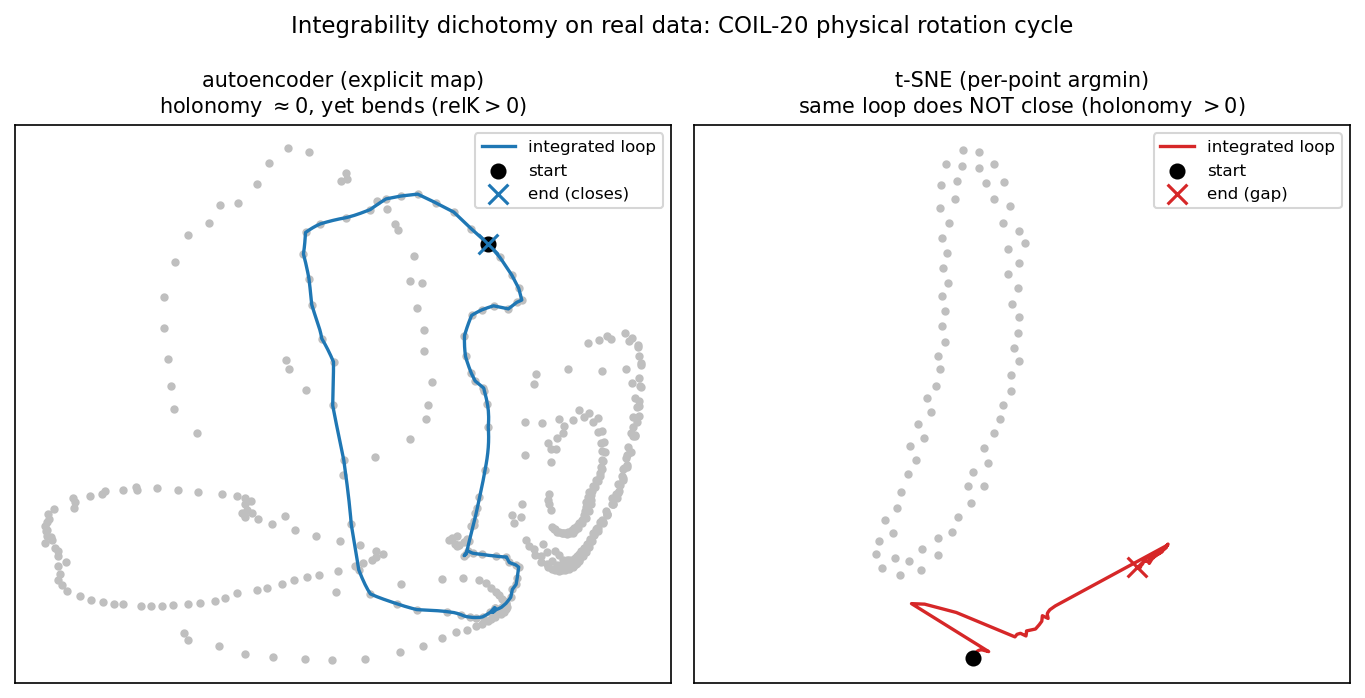}
\caption{The integrability dichotomy on a real physical loop (COIL-20 rotation cycle), both sides
integrated with the identical RK scheme. \emph{Left:} an autoencoder (explicit map $\Phi$), trained
on the other objects with the target held out, returns the target's rotation loop to its start at the
floating-point floor (holonomy $\approx0$) while still bending along the path ($\mathrm{relK}>0$).
\emph{Right:} the t-SNE per-point argmin flow of the \emph{same} held-out loop does not close:
zero against non-zero on identical data (magnitude not claimed; see text).}
\label{fig:coildich}
\end{figure*}

\smallskip\noindent\textbf{The integrability dichotomy on this real loop.}
The same rotation cycle also makes the cross-construction dichotomy of Sec.~\ref{sec:dichotomy}
concrete on real data (Fig.~\ref{fig:coildich}). We hold the target object out and read its rotation
loop two ways under the \emph{identical} Runge--Kutta scheme. An autoencoder trained on the remaining
objects (an explicit map $\Phi$, for which the loop is OOS just as it is for the argmin
side) integrates the loop back to its start at the floating-point floor ($\sim\!10^{-8}$ relative
to the embedding diameter), because $\omega=d\Phi$ is an exact form with no caustic; yet the encoder
still bends along the rotation path (its curvature relative to the transport is non-zero), so the
differential reading stays alive where the integral reading is vacuous. The t-SNE argmin flow of the
same loop, by contrast, does not close: zero against non-zero on the same real data, a
separation of many orders. The explicit side is structurally zero, so the argmin non-closure is
consistent with a branch obstruction; the explicit-map control integrates a
well-conditioned problem, so it does not by itself exclude integrator error on the ill-conditioned
one.

\begin{figure*}[t]\centering
\includegraphics[width=\textwidth]{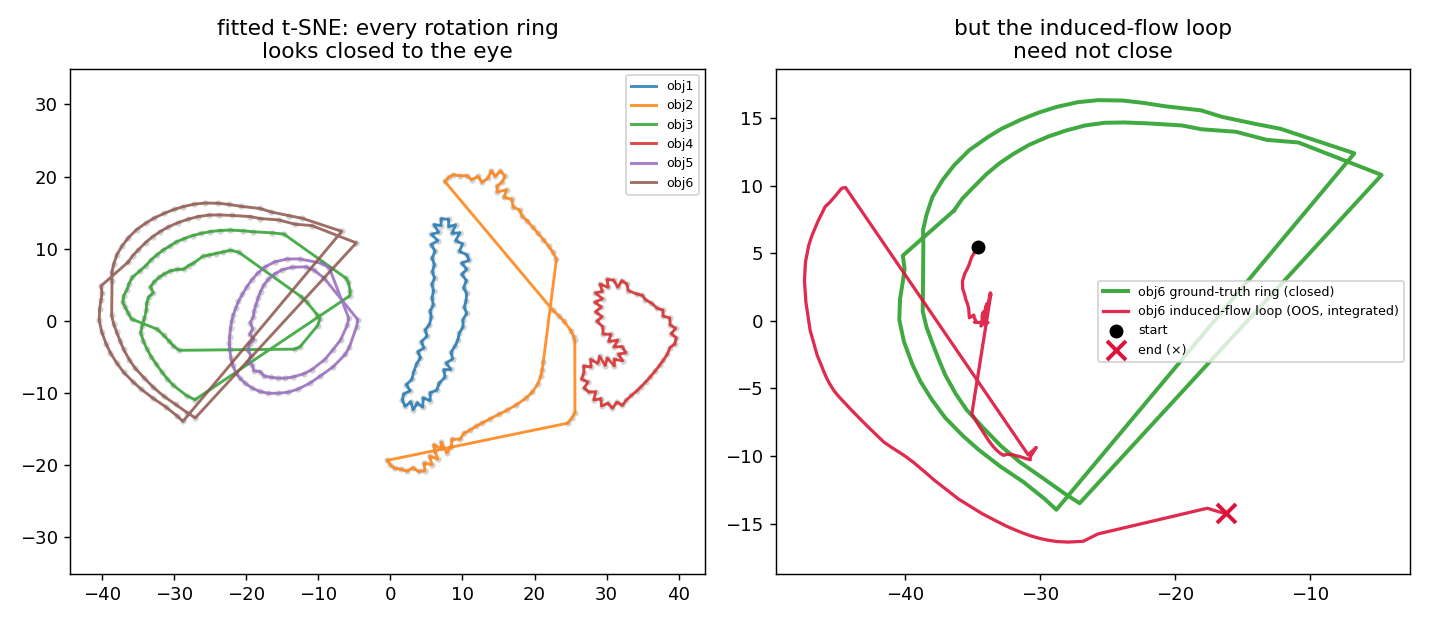}
\caption{COIL-20 rotation-cycle concept exhibit (qualitative). Every fitted ring looks equally closed
to the eye (left); integrating a held-out object's rotation loop through the induced flow illustrates
that the map need not return the closed high-d viewpoint cycle to itself (right: the
integrated loop against its ground-truth ring). Ground truth: every high-d rotation loop is
closed by construction, so any failure to close is a pure map artifact.}
\label{fig:coil}
\end{figure*}

\section{Inverse back-transport: the fold-and-hole testbed}\label{app:inverse}
This exhibit expands the empirical claim of Sec.~\ref{sec:inverse} (Proposition~P4). The main text
states the decomposition and its numbers inline; the figure shows the controlled testbed on which they
are measured. A symmetry-built dataset has three region types: \emph{reachable} (a single sheet),
\emph{fold} (two high-d sheets symmetric about the anchor plane share one induced placement,
$\Phi_+\!=\!\Phi_-$ to machine zero), and \emph{empty} (a punched hole). The two failure modes of the
inverse are certified by objects the framework already computes --- density for the empty region, the
\emph{forward} holonomy for the fold --- while a round-trip self-consistency residual detects neither,
confirming that the inverse needs no new certificate.

\begin{figure*}[t]\centering
\includegraphics[width=\textwidth]{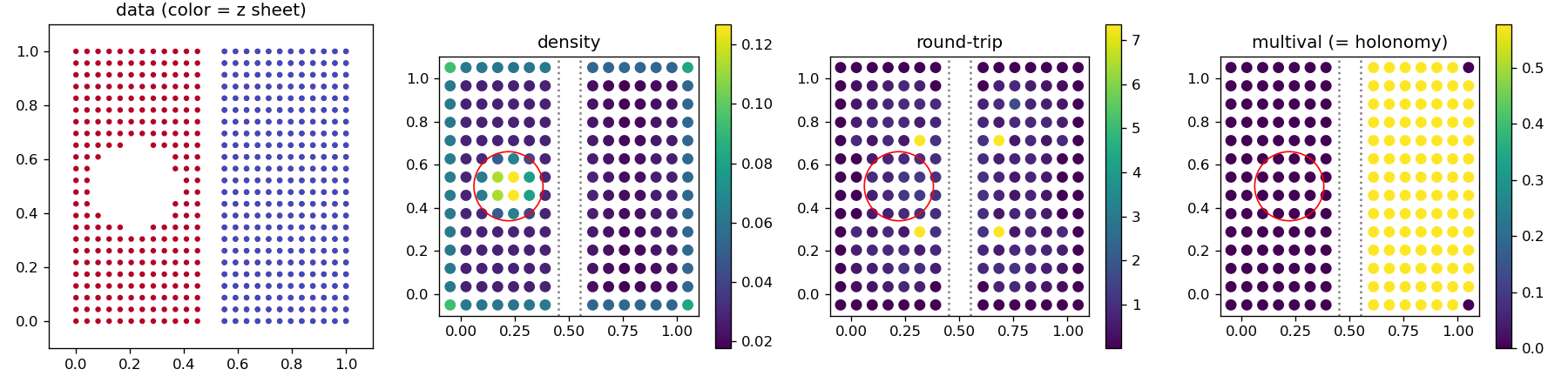}
\caption{\textbf{The inverse inherits the framework's own certificate.} The symmetry-built testbed
above: \emph{reachable} (single sheet), \emph{fold} (two sheets share one induced placement),
\emph{empty} (a punched hole). Across drawn query locations, a trivial \emph{density} check detects the
empty region (AUC $1.00$) but not the fold; the \emph{multivaluedness} of the induced transport --- the
forward holonomy --- detects the fold (AUC $1.00$) but not the empty region; a \emph{round-trip}
self-consistency residual detects neither (it extrapolates over the hole and returns one valid branch at
the fold; incremental AUC vs density $-0.15$, CI $[-0.28,-0.01]$). The inverse's reliability failure
decomposes as empty~(density)~$\cup$~fold~(holonomy).}
\label{fig:inverse-law}
\end{figure*}

\renewcommand{\refname}{Appendix References}
\putbib[refs]
\end{bibunit}   